%% file: main.tex
\documentclass[journal,applsci,article,submid,moreauthors,pdflatex]{Definitions/mdpi}
\firstpage{1} 
\pubvolume{xx}
\issuenum{x}
\articlenumber{x}
\pubyear{2019}
\copyrightyear{2019}
\history{Received: date; Accepted: date; Published: date}

\usepackage[utf8]{inputenc}
\usepackage{amsmath}
\usepackage{amssymb}
\usepackage{natbib}
\usepackage{graphicx}
\usepackage{todonotes}
\usepackage{url}
\usepackage{array}
\usepackage{multirow}
\usepackage{graphicx}
\usepackage{booktabs}
\usepackage{bm}
\usepackage{csquotes}
\usepackage{makecell}

\usepackage{amsfonts}
\usepackage{hyperref}
\usepackage{pifont}
\newcommand{\xmark}{\ding{55}}
\newcommand{\cmark}{\ding{51}}

\usepackage[TS1,T1]{fontenc}

\Title{OCR4all - An Open-Source Tool Providing a (Semi-)Automatic OCR Workflow for Historical Printings}
\Author{Christian Reul, Dennis Christ, Alexander Hartelt,\\Uwe Springmann, Christoph Wick, Christine Grundig,\\Andreas Büttner, and Frank Puppe}

\Author{Christian Reul $^{1}$, Dennis Christ $^{1}$, Alexander Hartelt $^{1}$, Nico Balbach $^{1}$, Maximilian Wehner $^{1}$, Uwe Springmann $^{2}$, Christoph Wick $^{1}$, Christine Grundig $^{3}$, Andreas Büttner $^{4}$, and Frank Puppe $^{1}$}

\AuthorNames{Christian Reul, Dennis Christ, Alexander Hartelt, Nico Balbach, Maximilian Wehner, Uwe Springmann, Christoph Wick, Christine Grundig, Andreas Büttner, and Frank Puppe}

\address{%
\\
$^{1}$ \quad Chair for Artificial Intelligence and Applied Computer Science, University of Würzburg, 97074 Würzburg, Germany
\\
$^{2}$ \quad Center for Information and Language Processing, LMU Munich,
80538 Munich, Germany
\\
$^{3}$ \quad Institute for Modern Art History, University of Zurich, 8006 Zurich, Switzerland
\\
$^{4}$ \quad Institute for Philosophy, University of Würzburg, 97074 Würzburg, Germany
\\
}

\corres{Correspondence: christian.reul@uni-wuerzburg.de}

\abstract{
Optical Character Recognition (OCR) on historical printings is a challenging task mainly due to the complexity of the layout and the highly variant typography.
Nevertheless, in the last few years great progress has been made in the area of historical OCR, resulting in several powerful open-source tools for preprocessing, layout recognition and segmentation, character recognition and post-processing.
The drawback of these tools often is their limited applicability by non-technical users like humanist scholars and in particular the combined use of several tools in a workflow. 
In this paper we present an open-source OCR software called OCR4all, which combines state-of-the-art OCR components and continuous model training into a comprehensive workflow.
A comfortable GUI allows error corrections not only in the final output, but already in early stages to minimize error propagations.
Further on, extensive configuration capabilities are provided to set the degree of automation of the workflow and to make adaptations to the carefully selected default parameters for specific printings, if necessary.
Experiments showed that users with minimal or no experience were able to capture the text of even the earliest printed books with manageable effort and great quality, achieving excellent character error rates (CERs) below 0.5\%.
The fully automated application on 19\textsuperscript{th} century novels showed that OCR4all can considerably outperform the commercial state-of-the-art tool ABBYY Finereader on moderate layouts if suitably pretrained mixed OCR models are available.
The architecture of OCR4all allows the easy integration (or substitution) of newly developed tools for its main components by standardized interfaces like PageXML, thus aiming at continual higher automation for historical printings. 
}


\begin{document}

\input{chapters/01_introduction.tex}
\input{chapters/02_related_work.tex}
\input{chapters/03_methods.tex}
\input{chapters/04_evaluations.tex}
\input{chapters/05_discussion.tex}
\input{chapters/06_conclusion.tex}

\authorcontributions{
Conceptualization: C.R. and F.P.;
Methodology: C.R. and U.S.;
Software: C.R., D.C., A.H., N.B., M.W., C.W., and A.B.; 
Formal Analysis: C.R., F.P., and C.W.;
Investigation, C.R., M.W., N.B., and F.P.;
Writing – Original Draft Preparation: C.R. and F.P.;
Writing – Review \& Editing: C.R., F.P., U.S., M.W., C.W., C.G., N.B., A.H., and A.B.;
Supervision: C.R., F.P., and U.S.;
Project Administration: C.R. and F.P.;
Funding Acquisition: F.P.
}

\funding{This research was funded by the Chair for Artificial Intelligence and the Centre for Philology and Digitality at the University of Würzburg as well as the Federal Ministry of Education and Research project \emph{Kallimachos}.}

\acknowledgments{The authors would like to express their gratitude to the entire \textit{Narragonien digital} workgroup around Brigitte Burrichter and Joachim Hamm who helped to established, test, and optimize the OCR4all workflow in close cooperation.
Furthermore, we would like to thank the \textit{Opera Camerarii} team around Thomas Baier, Marion Gindhart, Joachim Hamm, and Ulrich Schlegelmilch for providing a valuable and challenging use case and test object for OCR4all.
Finally, our gratitude is due to everyone who supported the planning, implementation, evaluation, distribution, and presentation of OCR4all including Sophia Beckenbauer, Kevin Chadbourne, Björn Eyselein, Yannik Herbst, Raphaëlle Jung, Tanja Kohl, Florian Langhanki, Maximilian Nöth, and Ronja Wegrath.
}

\conflictsofinterest{The authors declare no conflict of interest.} 

\abbreviations{The following abbreviations are used in this manuscript:\\

\noindent 
\begin{tabular}{@{}ll}
CC & Connected Component \\
CER & Character Error Rate \\
CNN & Convolutional Neural Network \\
CPU & Central Processing Unit \\
CTC & Connectionist Temporal Classification \\
GPU & Graphical Processing Unit \\
GT & Ground Truth \\
LSTM & Long Short-Term Memory \\
OCR & Optical Character Recognition \\
\end{tabular}}

\reftitle{References}
\bibliographystyle{plain}
\bibliography{references}
\end{document}

%% file: chapters/01_introduction.tex
\section{Introduction}

While Optical Character Recognition (OCR) is regularly considered to be a solved problem \cite{DoermannTombre2014}, gathering the textual content of historical printings using OCR can still be a very challenging and cumbersome task, due to various reasons.
Among the problems that need to be addressed for early printings is the often intricate layout containing images, artistic border elements and ornaments, marginal notes, and swash capitals at section beginnings whose positioning is often highly irregular.
The segmentation of text and non-text can currently not be done completely automatically to a high degree of accuracy.

Also, the non-standardized typography represents a big challenge for OCR approaches.
While modern fonts can be recognized with excellent accuracy by so-called omnifont or polyfont models, that means models pretrained on a large variety of customarily used fonts, the lack of computerized old fonts prevents the easy construction of such polyfont or mixed font models for old printing material and one needs to resort to individual model training instead. 

Therefore, very early printings like incunabula\footnote{Books printed before 1501.} but also handwritten texts usually require book-specific training in order to reach Character Error Rates (CERs) well below 10 or even 5\% as shown by Springmann et al.\ \cite{springmann2016automatic}\cite{springmannluedeling2017} (printings) and Fischer et al.\ \cite{fischer2009automatic} (manuscripts).
For a successful supervised training process ground truth (GT) in form of line images and their corresponding transcriptions has to be manually prepared as training examples.

Additionally, the highly variant historical spelling, including a frequent use of abbreviations, severely hinders automatic lexical correction, since sometimes identical words are spelled differently not only among different books of the same period but even within the same book.

In the last few years some progress has been made in the area of historical OCR, especially concerning the character recognition problem.
On the technical side an important milestone was the introduction of recurrent neural networks with Long Short Term Memory (LSTM) \cite{hochreiter1997long} trained using a Connectionist Temporal Classification (CTC, \cite{graves2006connectionist}) decoder which Breuel et al.\ applied to the task of OCR \cite{breuel2013lstm}.
The LSTM approach was later extended by deep convolutional neural networks, pushing the recognition accuracy even further \cite{breuel17hybrid,wick2018comparison}.

On the methodical side several improvements have been made by the introduction of voting ensembles, trained with a single OCR engine, whose results are suitably combined \cite{reul2018voting}, and by a pretraining approach which allows to use existing models instead of training from scratch \cite{reul2017transfer}.

The current paper describes our efforts to collect these recent advances into an easy to use software environment called OCR4all that runs as a docker image on several platforms (Linux, Mac, Windows) and enables an interested party to obtain a textual digital representation of the contents of these printings.
OCR4all covers all steps of an OCR workflow from preprocessing, document analysis (segmentation of text and non-text regions on a page), model training, to character recognition of the text regions.
Estimates of the remaining error rates are also given.
Work to include postcorrection methods to correct these residual errors is under way.
Our focus is throughout on an easy-to-use and efficient method, employing automatic methods where feasible and resorting to manual intervention where necessary.
In the following we give a short overview over the steps of a typical OCR workflow and how we address the challenges that arise for early printings.

\subsection{Steps of a Typical OCR Workflow}

\begin{figure}[!htbp]
    \centering
    \includegraphics[width=\linewidth]{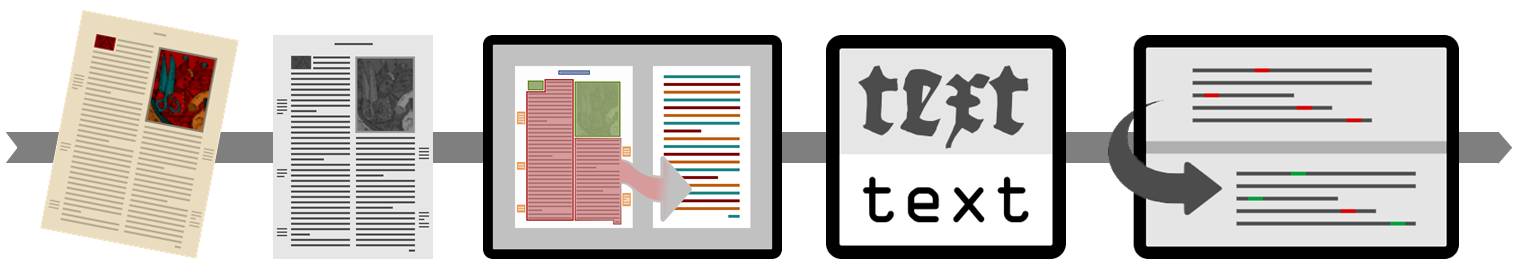}
    \caption{Main steps of a typical OCR workflow. From left to right: original image, preprocessing, segmentation, OCR, postcorrection.}
    \label{fig:schema}
\end{figure}

The character recognition in itself only represents one subtask within an OCR workflow, which usually consists of four main steps (see Figure \ref{fig:schema}) which often can be split up into further sub steps.
We use the term ``OCR'' as a separate main step within the OCR workflow as other notations like ``recognition'' would be misleading since the step comprises more sub task than the text recognition alone.

\begin{enumerate}

\item
\textbf{Preprocessing}: First of all, the input images have to be prepared for further processing.
Generally, this includes a step which simplifies the representation of the original color image by converting it into binary and sometimes grayscale, enabling further image processing operations later on, as well as a deskewing operation in order to get the pages into an upright position, which simplifies the  upcoming steps.
Additional routines like dewarping to rectify a distorted scan or denoising or despeckling to clean up the scan may be performed.
Beforehand, it can be worthwhile to crop the printing area in order to remove unwanted scan periphery.

\item
\textbf{Segmentation}: Next, one or several segmentation steps have to be conducted, mostly depending on the material at hand and the requirements of the user.
After separating the text regions from non-text areas individual text lines have to be identified and depending on the used OCR approach split up even further into single glyphs.
Optionally, non-text elements can be further classified (images, ornaments, ...), while text regions can be broken down into more or less fine-grained semantic classes (running text, headings, marginalia, ...), already on layout level.
Another important sub task is the determination of the reading order which defines the succession of text elements (region and/or lines) on a page.
Naturally, the option to manually correct the results achieved during these sub tasks is highly desired, preferably via a comfortably-to-use GUI.

\item
\textbf{OCR}: The recognition of the segmented lines (or single glyphs) leads to a textual representation of the printed input.
Depending on the material at hand and the user requirements this can either be performed by making use of existing models trained on a variety of fonts which somewhat fit the type in use, so-called mixed models, and/or by training models which are specifically geared to recognize the font it was trained on.
Again, for a comfortable correction of the textual OCR output and for producing good examples to be used for book-specific training a GUI is mandatory.

\item
\textbf{Postprocessing}: The raw OCR output can be further improved during a postprocessing step, for example by incorporating dictionaries or language models.
This step can be combined with the manual correction which would usually take place after the automatic postprocessing. 

\end{enumerate}

As for the final output, plain text, that is the (postprocessed) OCR output, has to be considered the minimal solution.
Apart from that, almost all of the information acquired during the entire workflow can be incorporated into the final output: region coordinates and their types, line coordinates, character positions and information about how sure the OCR is about its predictions, and others.
Several formats which can incorporate most or all of the aforementioned information have been proposed, for example ALTO\footnote{\url{https://www.loc.gov/standards/alto/}}, hOCR \cite{breuel2007hocr}, or PAGE \cite{pletschacher2010page}.

Regarding the text recognition step it is worth mentioning that the models used for the recognition
can be obtained in different ways with highly varying degrees of effort based on the material at hand.
As quoted above modern fonts can usually be recognized by applying an existing standard model which has been trained on a variety of similar fonts, for example 19\textsuperscript{th} century Fraktur, due to the comparatively high degree of homogeneity of the typography.
On the contrary, (very) early prints often require type-specific training to reach character recognition rates in the high nineties.

\subsection{Challenges for the Users}
To produce training data for the OCR one has to manually transcribe text lines (considering a line-based approach), which is a highly non-trivial task when dealing with very old fonts which are often difficult to decipher and contain numerous ligatures and abbreviations whose transcription and/or decomposition requires knowledge about the historical language and the content of the texts.

In fact, this step and often also several other steps of the OCR workflow cannot be performed fully automatically and require the user to interact and to invest manual work.
The combination of all steps represents a highly interdisciplinary task and therefore requires both domain expertise regarding the content as well as technical expertise, a combination which is difficult to come by in a single person.
Since it is not possible to simplify the content related part of the problem, we have to focus on the technical aspect.
Fortunately, large parts of these steps can be covered by open-source tools such as OCRopus, Tesseract or Calamari which have been made available as open source.

While these tools are highly functional and very powerful their usage can be quite complicated, as they

\begin{itemize}
    \item in most cases lack a comfortable GUI which leaves the users with the often unfamiliar command line usage
    \item usually rely on different input/output formats which requires the users to invest additional effort in order to put together an end-to-end OCR workflow
    \item sometimes require complicated and error prone installation and configuration procedures where, e.g., the users have to deal with missing dependencies
    \item have a steep learning curve (at least for non-technical users)
\end{itemize}

These aspects are particularly problematic for inexperienced users with limited technical background.

Unfortunately, this often includes
humanities scholars 
as one of the main target audiences for all tools which allow to produce machine actionable text from scans of historical printings.
Making an entire workflow available to and usable by non-technical users is a challenging task, since most tools usually do not cover the entire workflow described above, at least not in a satisfactory manner, but rather excel on smaller sub tasks.
Combined with the shortage of (user friendly and GUI supported) ways to manually interfere with the process and turn it into a semi-automatic approach, this considerably reduces the applicability of existing open-source tools.

\subsection{OCR4all}

To deal with these issues, we present our open-source tool OCR4all\footnote{\url{https://www.uni-wuerzburg.de/en/zpd/ocr4all}}\footnote{\url{https://github.com/OCR4all}} which aims to encapsulate a comprehensive OCR workflow into a single Docker\footnote{\url{https://www.docker.com/}} application, ensuring easy installation and platform independency.
The goal is to make the capabilities of state of the art tools like OCRopus or Calamari available within a comprehensible and applicable semi-automatic workflow to basically any given user.
This is achieved by supplying the users with a comfortable and easy to use GUI and a modular approach, allowing for an efficient correction process in between the various workflow steps in order to minimize the negative effects of consequential errors.
Another important aspect is the option to iteratively reduce the CER by constantly retraining the recognition models on additional training data which has been created during the processing of a given book.
During development the primary goal was to identify a workflow and tool composition which enables the users to deal with even the earliest printed books on their own and extract their textual content with great quality.
Due to the challenges concerning condition, layout and typography described above, this is far from a trivial task and often requires the users to invest a substantial amount of manual effort into correcting segmentation results and transcribing line images in order to produce training data for the book specific models.
However, in our experience, humanist scholars who are dealing with early printed books are usually perfectly fine with investing the required effort to obtain high quality OCR results which had been considered almost impossible to achieve on this material only a decade ago \cite{rydberg2009digitizing}.
This is especially true since the alternatives are either to manually transcribe everything or to not get to the text at all.
Naturally, we did not want to restrict the users to the processing of very early printed books and therefore added further functionality to ensure a fluent passage towards a fully automated approach when dealing with later and more uniform works.

Despite our focus on user friendliness, operating OCR4all is still not an entirely trivial task (and will not be for the foreseeable future), especially when dealing with very early prints with a complex and irregular layout as well as a non standard typeface that makes a thorough book-specific training indispensable.
Consequently, there is an obvious need for detailed, comprehensible, and descriptive operating instructions.
To start things off we provided both, a setup guide and a comprehensive step by step user manual together with some example data at GitHub\footnote{\url{https://github.com/OCR4all/getting_started}} and set up a mailing list\footnote{\url{https://lists.uni-wuerzburg.de/mailman/listinfo/ocr4all}} where we inform about latest developments and new version releases.

\subsection{Outline of this Work}

The paper is structured as follows:
First, section \ref{sec:RW} provides an overview over important contributions concerning OCR relevant to our task.
In section \ref{sec:ocr4all} we thoroughly describe OCR4all including the overall workflow and the single sub modules.
Next, we perform several experiments on a variety of historical printings.
The results are discussed in section \ref{sec:discussion} before section \ref{sec:conclusion} concludes the paper by summing up the insights and pointing out our goals for the future of OCR4all.

%% file: chapters/02_related_work.tex
\section{Related Work}
\label{sec:RW}

In this section we give a comprehensive overview over OCR related tools and topics.
First, we focus exclusively on the OCR since it has to be considered the core task of the entire workflow.
Afterwards, we discuss further steps and tools which (aim to) provide an entire OCR workflow.
Finally, we introduce additional notable OCR related tools and projects.

\subsection{Optical Character Recognition}
After introducing the historical development in the area of OCR we briefly discuss advantages and disadvantages of mixed models compared to book-specific training and finally highlight the recognition capabilities of several available OCR engines.

\subsubsection{Historical Development and State of the Art}
Text recognition can be considered as one of the earliest computer vision tasks \cite{Schantz1982}.
Therefore, we begin our survey of related work with a short sketch of the general recognition approaches.

\paragraph{Glyph-based Recognition}
For a long time segmenting printed texts into single glyphs which are then classified individually was considered the go-to OCR approach.
After identifying a glyph a feature extraction step takes place before the gathered information is used to assign a character class.
This approach was used by all available OCR engines until very recently, e.g. by the open source OCR engine Tesseract\footnote{\url{https://github.com/tesseract-ocr/tesseract}} before version 4.0.

The main drawback of this method is the need to precisely identify every single glyph which can be a very challenging task especially when dealing with older printings where the segmentation step leads to either splits or merges of glyphs on a regular basis, as the glyph contours have lost their uniform ink impression and get segmented as individual pieces, or contours of neighboring glyphs have become fuzzy  and tend to touch each other leading to segments containing several individual glyphs that cannot subsequently be classified.
Furthermore, creating training data for training a recognition model based on real printings (as opposed to train on synthetical images from existing computer fonts) is a cumbersome and time consuming task. Still, Kirchner et al. \cite{kirchner2016ocr} showed that it is basically possible to train book- or rather type-specific models with Tesseract 3 using Aletheia and Franken+\footnote{\url{https://emop.tamu.edu/outcomes/Franken-Plus}}.
After manually identifying examples for each glyph class Franken+ supported the creation of the required Tesseract 3 training format.
A model trained on an incunabulum was then applied to other books of the same print shop using the exact same type and resulted in CERs between 4 and 8\%. However, due to the high amount of manual work required to produce such a model this approach seemed only practicable if one desires to capture a variety of works printed with the same letter types.

\paragraph{Line-based Recognition using LSTM Networks}
In 2013 Breuel et al.\ introduced a segmentation free approach (no segmentation beyond the line level) in their ground-breaking paper \cite{breuel2013lstm} which is capable to recognize entire text line images at once.
This is possible by utilizing recurrent neural networks with an LSTM architecture trained using the CTC algorithm.
After resizing a line image to a fixed height the image is cut into vertical stripes with a width of one pixel.
The pixel values of these stripes (usually binary or grayscale) are fed into the neural network which produces a probability distribution over the entire glyph alphabet for each stripe, usually by processing the input sequence two times: from left to right and from right to left (bidirectional LSTM).
Finally, the output sequence is generated by applying a CTC decoder.

The line-based OCR approach was not only shown to outperform the glyph-based approach considerably, but also offers the advantage of a much easier GT production and training process.
Lines chosen for training can simply be transcribed as a whole since a line image and the corresponding transcription completely suffice to serve as a training example without the need for any further information about glyph positions or bounding boxes.
Those improvements also enabled an efficient high quality processing of even the earliest printed books as shown by Springmann et al.\ with CERs from individually trained models of the order of 2\% \cite{springmann2016automatic, springmannluedeling2017}.

\paragraph{Line-based Recognition using CNN/LSTM-Hybrid Networks}
A further refinement of the LSTM approach was introduced in 2017 by Breuel \cite{breuel17hybrid} who added Convolutional Neural Networks (CNNs), which showed to be very effective in a variety of image processing task \cite{mane2017survey}, as additional layers in front of the LSTM.
Each CNN performs a convolution of the original line image using different filters whose parameters are learned during the training process producing a feature map that highlights the most descriptive parts of the input image.
After a pooling operation the resulting images are then either passed into another CNN or vertically concatenated and passed into the LSTM layer. 

\begin{figure}[!htb]
    \centering
    \includegraphics[width=0.25\linewidth]{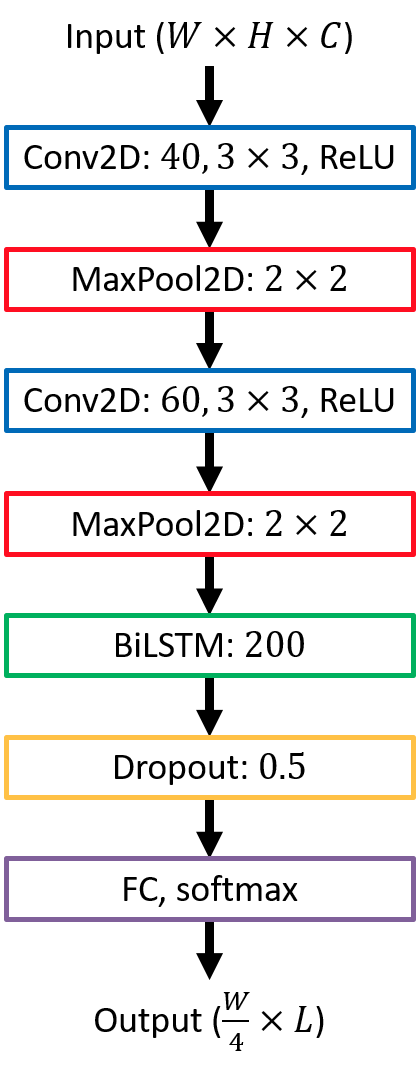}
    \caption{Calamari's default network architecture:
To begin with, the input image with width \textit{W}, height \textit{H}, and \textit{C} color channels is passed into the first convolutional layer.
Applying 40 filter operations with a kernel size of $3\times3$ results in 40 feature maps which are then reduced by a max pooling operation with a pool size of $2\times2$.
The two steps are then repeated but this time 60 features are used during the convolution.
Next, the feature maps are vertically concatenated and passed into a bidirectional LSTM with 200 hidden nodes.
A dropout layer with a dropout rate of 0.5 is introduced to reduce the effect of overfitting.
Finally, for every horizontal position a fully connected layer with softmax leads to the final output probability matrix with $\frac{W}{4}$ columns (the original width of the input image is reduced by factor 4 because of the two pooling operations) and \textit{L} rows, with \textit{L} representing the number of labels, which is the alphabet size plus the blank label.}
    \label{fig:calamari_architecture}
\end{figure}

Fig. \ref{fig:calamari_architecture} shows an exemplary network architecture.

This CNN/LSTM-hybrid method has shown to be very successful in various application scenarios and therefore also represents the current state-of-the-art of modern OCR engines like Calamari, Tesseract (since version 4.0) and OCRopus 3.

\paragraph{Calamari}
We chose Calamari \cite{wick2018calamari} as our OCR engine since it is available under an open-source license and previous tests have demonstrated its advantages compared to other OCR engines regarding recognition capabilities and speed\cite{wick2018comparison}.
It focuses solely on the OCR training and recognition step and does not offer any preprocessing, segmentation or postprocessing capabilities, which is why it will not be covered in the upcoming comprehensive discussion of tools that provide a full OCR workflow.
It implements a deep CNN-LSTM network architecture and its Tensorflow backend enables GPU support for very fast training and recognition.

Furthermore, Calamari natively supports several techniques, resulting in higher recognition rates which we will briefly explain in the following.
First, a \emph{cross-fold training procedure with subsequent confidence voting} in order to reduce the CER on early printed books was implemented \cite{reul2018voting}.
By dividing the GT in $N$ different folds and aligning them in a certain way, it is possible to train $N$ strong but also diverse models which act as voters in a newly created confidence voting scheme.
Second, the so-called \emph{pretraining} functionality allows to build from an already available Calamari model instead of starting training from scratch which not only speeds up the training process considerably but also improves the recognition accuracy \cite{reul2017transfer}.
Third, \emph{data augmentation} using the routines of \emph{ocrodeg}\footnote{\url{https://github.com/NVlabs/ocrodeg}} for generating noisy variations of training material.
All three techniques, pretraining, voting, and data augmentation, were included into Calamari \cite{wick2018comparison}.

Finally, Calamari provides several interfaces for more complex data representations than image/text pairs on line level, most notably PageXML.
Combined with the highly modular structure this ensures a straight forward integration into existing and future OCR workflows.

\subsubsection{Mixed Models}
While the best results can usually only be achieved by training book-specific models the out-of-the-box application of existing mixed models represents a baseline option that may already be good enough and could be done automatically.
However, even a fully automated approach becomes unattractive if it cannot yield satisfactory OCR results.
Consequently, the applicability of mixed models considerably depends on two factors: the use case and the material at hand.
In fact, the quality of an OCR text directly influences the possible areas of application, for example while critical editions aim for perfection, tasks like key word search can usually even deal with heavily flawed texts.
Regarding the to-be-processed material it is apparent that mixed models can deal best with works which are printed in a similar type as the training material on which the models have been trained.
Since the variance between printing types started off very high in the incunabula age and decreased over the centuries, it is clear that training wide applicable mixed models becomes the more challenging the older the target material gets.
This effect as well as a general comparison of the effectiveness of mixed and book-specific models was investigated by Springmann et al.\ \cite{springmann2016automatic, springmannluedeling2017}) who relied on OCRopus 1 as their OCR engine:

First, they performed experiments on a corpus consisting of twelve books printed with Antiqua types between 1471 and 1686 with a focus (ten out of twelve) on early works produced before 1600.
After dividing the corpus into two distinct sets of six books each a mixed model was trained on both of them.
The evaluation of each model on the respective held-out books yielded an average CER of 8.3\% with the individual CERs ranging from 21\% to below 2\%.
Only two books scored a CER higher than 10\%, both of them incunabula.
As expected, training book-specific models and evaluating them on held-out data of the same book resulted in considerably better recognition results ranging from 5.3\% to below 1\% and an average of 2.2\%.

Second, a similar experiment was conducted as part of a case study on the RIDGES corpus\footnote{\url{http://korpling.org/ridges}, see \cite{odebrecht2017ridges}} consisting of 20 German books printed between 1487 and 1870 in Fraktur. 
After applying the same methodology as mentioned above the mixed models scored an impressive average CER of 5.0\% with individual results ranging from 17\% to below 1\%.
Similar to the first case study the two oldest books performed worst with CERs over 10\%.
As a matter of fact, the individually trained models again performed considerably better, reaching an average CER of around 2.1\% with the worst book still achieving 5.4\%.

In both case studies the oldest printings proved to be the most difficult ones for the mixed models to recognize, as expected.
This highlights the general difficulty of using mixed models: To yield high quality results the types at hand have to fit (parts of) the material the model was trained on as much as possible.
While this becomes more likely when adding more works to the training set, a fit can never be guaranteed.



\subsubsection{Evaluation of OCR Engines}

In the following we compare existing OCR engines and identify missing features which we later address with our own solution based upon Calamari (recognition accuracy and speed), mixed models, training on historical data, and the necessary tooling for a complete workflow.

\paragraph{Book-specific Training on Early Printed Books}
A thorough comparison of a shallow LSTM (OCRopus 1) and a deep CNN/LSTM hybrid (Calamari) is given in \cite{wick2018comparison}.
Three early printed books, printed between 1476 and 1505 in German and Latin, were used as training and evaluation data.
Book-specific models were trained using 60, 100, 150, 250, 500 and 1,000 training lines.
The results showed, as anticipated, that the advantage of the deep network grew with an increasing number of lines used for training, yielding an average improvement in CER of 29\% for 60 lines and 43\% for 1,000 lines.
As for the training and prediction times the deep network, despite having a considerably more complex network structure and more trainable parameters resulting in a higher number of necessary operations, outperformed the shallow one when using four CPU threads or more.
The reason for this are the pooling layers which reduce the dimensions of the image by a factor of 4 leading to a considerably speedup during the expensive LSTM and CTC operations.
Running the training and prediction on a GPU led to a further speedup by a factor of at least six and four, respectively.

\paragraph{Modern English and 19th Century Fraktur}
In \cite{wick2018calamari} Wick et al.\ compared Calamari, OCRopus 1, OCRopus 3, and Tesseract 4 on two public datasets:
First, the UW3\footnote{University of Washington Database III: \url{http://www.tmbdev.net/ocrdata-split}} dataset consisting of ca. 80,000 lines (70,000 for training and 10,000 for evaluation) from English speaking scientific publications from the late 20\textsuperscript{th} century printed in Antiqua.
Second, the DTA19 dataset which is a part of the GT4HistOCR corpus\footnote{\url{https://zenodo.org/record/1344132}} \cite{springmann2019ground} containing 39 German novels printed in Fraktur during the 19\textsuperscript{th} century.

Again, book- or rather corpus-specific training was performed with each engine using the respective default parameters. 
On the UW3 dataset Calamari achieved a CER of 0.155\%, considerably outperforming OCRopus 1 (0.870\%), OCRopus 3 (0.436\%), and Tesseract 4 (0.397\%).
The inclusion of confidence voting improved Calamari's result by another 26\% to a CER of just 0.114\%.

Evaluations on the DTA19 dataset led to similar observations with Calamari reaching a CER of 0.221\% (0.184\% with voting) compared to the significantly higher 1.59\% of OCRopus 1 and 0.907\% of OCRopus 3.
On this dataset no Tesseract 4 results were reported.

Regarding speed the average time needed to train or to predict a single line of the UW3 dataset was measured and compared.
When using a GPU Calamari required  8 ms to train and 3 ms to predict a line which proved to be considerably faster than OCRopus 3 (10 ms and 7 ms), while OCRopus 1 (850 ms and 330 ms), and Tesseract 4 (1,200 ms and 550 ms) are far behind due to their lack of GPU support.

\paragraph{Case Study on 19\textsuperscript{th} Century Fraktur Mixed Models}
A detailed comparison of our own mixed models with the historical Fraktur module of ABBYY Finereader on different textual material has not yet been undertaken.
However, a case study \cite{reul2019state} dealing with German 19\textsuperscript{th} century Fraktur scripts from various source materials (mostly novels but also a journal, different volumes of a newspaper as well as a dictionary) was performed.
Mixed models for Calamari and OCRopus 1 were produced by training on a very extensive corpus of German Fraktur data from the 19\textsuperscript{th} century which was completely distinct from the material of the evaluation set.

Altogether, ABBYY achieved a CER of 2.80\% but got significantly outperformed by Calamari (0.61\%) and even OCRopus 1 (1.90\%).
This suggests that, while the available pretrained ABBYY OCR and language models can achieve impressive results on material from the 19\textsuperscript{th} century, especially on high quality scans with a clear print image, the raw recognition capability cannot keep up with mixed models accurately trained with existing open-source OCR engines.
This gap becomes even wider when dealing with considerably older printings, on which ABBYY usually produces highly erroneous and therefore unusable output. This is doubtless due to the fact that the ABBYY models have not been trained on material from these periods.

\paragraph{Conclusion}
Based on the results presented above and our personal experience Table \ref{tab:ocr_comp} sums up and rates the capabilities of the most important available OCR engines in terms of historical OCR.

\begin{table}[!htbp]
\centering
\caption{Comparison and rating of the capabilities of four modern OCR engines.}
\label{tab:ocr_comp}
\begin{tabular}{c|cccc}
\toprule
\textbf{Step} & \textbf{ABBYY} & \textbf{OCRopus 3} & \textbf{Tesseract 4} & \textbf{Calamari} \\ 
\midrule

Recognition & \cmark\cmark & \cmark\cmark & \cmark\cmark & \cmark\cmark\cmark  \\
Training & \cmark &  \cmark\cmark & \cmark & \cmark\cmark\cmark  \\
Manual Correction & \cmark\cmark\cmark &  \xmark & \xmark  & \xmark \\

\bottomrule
\end{tabular}
\end{table}

As for the recognition the main criteria are accuracy and speed.
Since we consider postprocessing using dictionaries and language models to be an individual step in the workflow, we rate the raw recognition capabilities of the engines.
Due to the results presented above, the best rating goes to Calamari.

Regarding the training, we rate the engines mainly based on their speed and effectiveness but also take into account the user friendliness when it comes to training on real data.
OCRopus 3, Tesseract 4, and Calamari in general allow pairs of line images and their transcriptions as training input which is very comfortable and straight forward for the user.
While Calamari can deal with the image/text pairs directly, just like OCRopus 1, OCRopus 3 requires to create a .tar file comprising the data.
As for Tesseract 4, the training of models on real historical data has been considered at least impracticable for several years until recently a solution was discovered and made publicly available\footnote{\url{https://github.com/OCR-D/ocrd-train}}.
However, this requires an extension to the standard training tools.
While it is basically possible to train single glyphs and consequently a book-specific model using ABBYY, this is a tedious and ineffective task which seems to be mainly geared towards the recognition of quite specific ornament letters. This effectively limits the recognition capability to the expensive existing historical models one has to licence from ABBYY.

ABBYY offers a comprehensive set of support tools for the manual postcorrection including a synoptic image/text view, markers for possible errors based on recognition confidence and dictionaries, and a selection of possible alternatives.
While OCRopus 1 at least allows to create a browser based synoptic view,
OCRopus 3, Calamari, and Tesseract 4 do not offer any form of user interaction regarding the correction of the OCR output.

\subsection{Tools Providing an OCR Workflow}
Before discussing the various OCR workflow tools in detail we first give an overview of their respective capabilities in Table \ref{tab:workflow_tools}.
Regarding he OCR step we mostly incorporated the ratings from Table \ref{tab:ocr_comp} since, as shown above, there are several detailed comparative evaluations available which the other steps are lacking.
As Calamari represents our main OCR engine we adopted its ratings. 
There is a single exception: since OCR4all offers a line-based synoptic correction view including some user conveniences like a customizable virtual keyboard but currently lacking a dictionary or confidence based error detector, we adjusted the rating for \textit{manual correction} accordingly.

In the following, we discuss the four tools from Table \ref{tab:workflow_tools} but also briefly introduce other workflow tools as well as projects that deal with the OCR workflow.

\begin{table}[!htbp]
\centering
\caption{
Comparison of existing tools providing an OCR workflow with OCR4all.
}
\label{tab:workflow_tools}
\begin{tabular}{cc|ccc|c}
\toprule

\textbf{Step} & \textbf{Sub Task} & \textbf{ABBYY} & \textbf{OCRopus 3} & \textbf{Tesseract 4} & \textbf{OCR4all} \\ 

\midrule

\multirow{2}{*}{Preprocessing} 
& Deskewing & \cmark & \cmark & \cmark & \cmark  \\
& Binarization & \cmark & \cmark & \cmark & \cmark  \\

\midrule

\multirow{5}{*}{Segmentation} 
& Image/Text & \cmark & \xmark & \cmark & \cmark  \\
& Semantic Distinction & \xmark & \xmark & \xmark & \cmark  \\
& Line Segmentation & \cmark & \cmark & \cmark & \cmark  \\
& Reading Order & \cmark & \cmark & \cmark & \cmark  \\
& Manual Correction & \cmark & \xmark & \xmark & \cmark  \\

\midrule

\multirow{3}{*}{Historical OCR}
& Recognition & \cmark\cmark & \cmark\cmark & \cmark\cmark & \cmark\cmark\cmark  \\
& Training & \cmark &  \cmark\cmark & \cmark & \cmark\cmark\cmark  \\
& Manual Correction & \cmark\cmark\cmark &  \xmark & \xmark  & \cmark\cmark \\

\midrule

\multirow{2}{*}{Postprocessing} 
& Dictionaries & \cmark & \xmark & \cmark & \xmark  \\
& Language Modelling & \cmark & \xmark & \cmark & \xmark  \\

\midrule
\midrule
open-source & - & \xmark & \cmark & \cmark & \cmark \\

\bottomrule
\end{tabular}
\end{table}

\subsubsection{ABBYY}
At least on contemporary material the proprietary ABBYY OCR engine\footnote{\url{https://www.abbyy.com}} clearly defines the state of the art for preprocessing, layout analysis, and OCR.
A wide variety of documents with considerably differing layouts can be processed by the fully automated segmentation functionality whose results can be manually corrected, if necessary.

Especially regarding the character recognition ABBYY's focus clearly lies on modern printings since this represents their bulk business.
Currently (July 2019), their products support close to 200 recognition languages offering strong language models and dictionary assistance for about a quarter of them.
Despite the focus on modern prints the repertoire also includes the recognition of historical European documents and books printed in six languages.

Apart from its closed source and proprietary nature ABBYY's shortcomings in the area of OCR of (very) early printings lead to the conclusion that it does not fit the bill despite its comprehensive and powerful preprocessing, segmentation, and recognition capabilities (on later material) as well as its easy setup and comfortable GUI.

\subsubsection{Tesseract}
Just like ABBYY the open-source OCR engine Tesseract provides a full OCR workflow including built-in routines for preprocessing like deskewing and binarisation as well as for layout analysis, but overall it is significantly less successful than ABBYY.

Tesseract's OCR training and recognition capability recently (version 4.0+) have improved considerably due to the addition of a new OCR engine based on LSTM neural networks which clearly outperformed the character based approach during project internal experiments.
The old glyph-based recognition method is still supported and mixed models trained for both recognition approaches and a wide variety of languages and scripts are openly available at the project's Github repository.
Similar to ABBYY and contrary to OCRopus 1/2/3 and Calamari, Tesseract supports the use of dictionaries and language modelling.
While Tesseract has its strengths in the fully automatic out of the box processing of modern texts it falls short when it comes to historical material.


\subsubsection{OCRopus}
The open-source toolbox OCRopus 1\footnote{\url{https://github.com/tmbdev/ocropy}} \cite{breuel2008ocropus, breuel2009recent, breuel2013lstm} comprises several Python-based tools for document analysis and recognition.
This includes highly performant algorithms for deskewing and binarisation as well as a segmentation module which extracts text lines from a page in reading order.
While the segmentation can quite comfortably deal with modern standard layouts, that means text-only pages with clearly separable columns, it tends to struggle with typical historical layouts with marginalia, swash capitals, etc.
When a page has already been split up into regions however, the line segmentation usually identifies the single lines very reliably and accurately at least when working with Latin script.
This is not a trivial task since in historical printings the letters of adjacent lines can severely overlap vertically or even touch each other.

OCRopus 1 was the first OCR engine to implement the pioneering line based approach for character recognition introduced by Breuel et al.\ \cite{breuel2013lstm} using bidirectional LSTM networks which allowed for considerably superior recognition capabilities compared to glyph-based approaches.
Furthermore, this method significantly simplified the process of training new models since the user just has to provide image/text pairs on line level, which can be created by using a html-based transcription interface in a browser.

While the general line-based recognition still defines the state of the art the shallow network structure consisting of just a single hidden layer has to be considered outdated by now.
The superiority of deeper architectures relying on a combination of CNN and LSTM layers has been shown well enough on different materials.
Nevertheless, OCRopus 1 still proves to be a cornerstone for OCR workflows dealing with historical printings mainly for two reasons.
First, the preprocessing usually achieves excellent results due to its robust deskewing approach as well as its adaptive thresholding technique used for preprocessing \cite{afzal2013robust} which can comfortably deal with pages even if they are in questionable condition.
Second, due to the robust line segmentation described above.

After the comparatively disregarded OCRopus 2\footnote{\url{https://github.com/tmbdev/ocropy2}} the third edition OCRopus 3\footnote{\url{https://github.com/NVlabs/ocropus3}} was released in May 2018.
It introduced a PyTorch-backend which enabled the utilization of deep network structures and GPU support, resulting in better recognition rates and faster training and prediction.
In the comparative study mentioned earlier\cite{wick2018comparison} OCRopus 3 achieved recognition results similar to Tesseract 4 while being significantly faster but was nevertheless considerably outperformed by Calamari.
Concerning the other steps in the OCR workflow like binarisation, deskewing, and segmentation OCRopus 3 almost exclusively relies on deep learning techniques.
To the best of our knowledge there are not yet any comparisons available between the traditional methods of OCRopus 1 and the new approach by OCRopus 3.

\subsubsection{Kraken}
The open-source OCR software Kraken\footnote{\url{https://github.com/mittagessen/kraken}} (see \cite{kiessling2017kraken} for the initial paper) is originally based on an OCRopus 1 fork and has been significantly cleaned up as well as extended since.
For example, Kraken seems to focus on the processing of Arabic text, 
resulting in an optimized line segmentation procedure which can deal with the specifics of Arabic script and a right-to-left text recognition support.
The underlying OCRopus 1 architecture was extended by a PyTorch backend enabling the training of deep networks consisting of a combination of CNN and LSTM layers.

In a very recent (July 2019) second publication \cite{kiessling2019kraken} the present state of the software is briefly introduced.
Apart from extensive recognition features like the support of right-to-left, bidirectional, and vertical writing, combined script detection and multiscript recognition are addressed.
Moreover, a trainable deep learning line extractor is currently being implemented to allow dealing with the highly variant challenges of different scripts when it comes to line segmentation.
Finally, results achieved on several publicly available data sets including historical ones are presented, mostly achieving above 98\% or even 99\% character accuracy.
Unfortunately, neither details about the training and evaluation procedure nor a comparison with other open-source OCR engines are provided.

Despite the application to historical data the focus of the engine seems to lie on more recent printings.
This is indicated by the fact that the results of the line segmentation are output as JSON files simply containing the line bounding boxes as straight rectangles.
As mentioned above, this can be quite problematic when working on earlier printings, especially when considering that Kraken also got rid of OCRopus 1's deskewing functionality.
Because of these shortcomings or rather these intended simplifications Kraken does not seem ideally suited for an application to historical printings.

\subsubsection{Transkribus}
A very comprehensive platform specialized on Handwritten Text Recognition (HTR) was developed within the Transkribus project\footnote{\url{https://transkribus.eu}} \cite{transkribus} which provides a web service to store and share documents or perform layout analysis, recognition, and training tasks on the server.
The main user interface is available as an open-source Java desktop application which allows the user to perform manual segmentation tasks or produce GT by transcribing lines.
Unfortunately, large parts of the software are not open-source, preventing the users from adapting or extending the code and from running the advanced recognition tools on their own hardware.
On 1\textsuperscript{st} of July 2019 a fee-based cooperative was founded that 
\enquote{serves as the basis for sustaining and further developing the Transkribus platform and related services and tools}.\footnote{\url{https://read.transkribus.eu/coop/}}

The technical partner for the development of the layout analysis and training and recognition software is the CITlab\footnote{\url{https://www.mathematik.uni-rostock.de/forschung/projekte/CITlab}} team at the University of Rostock whose approach performed best on the sub task of the detection of baselines, that is the line supporting the main bodies of characters within a text line, at a competition layout analysis for challenging medieval manuscripts at ICDAR2017 \cite{simistira2017icdar2017}.
Several related publications are available (see for example \cite{gruening2017robust,gruning2018two} for layout analysis and \cite{michael2019seq2seq} for HTR) but to the best of our knowledge the exact state of the software actually incorporated in Transkribus is not publicly known.
Therefore, the best source for results seems a recently (May 2019) published  talk\footnote{\url{https://www.slideshare.net/ETH-Bibliothek/transkribus-eine-forschungsplattform-fr-die-automatisierte-digitalisierung-erkennung-und-suche-in-historischen-dokumenten}} which biefly sums up some evaluations:
After training on close to 36,000 words corresponding to 182 pages a CER of 3.1\% and a WER of 13.1\% was achieved on a dataset from the 18\textsuperscript{th} century written by a single writer in German.
For Latin and French medieval material from many different writers the system scored a CER of 6.4\% and a WER of 22.1\% after being thoroughly trained on over half a million words corresponding to close to 1,200 pages of GT.
The application to printed text, more precisely to newspapers from the 18\textsuperscript{th} century, led to a CER of 0.81\% and a WER of 3.02\% was achieved after training on 180,000 words corresponding to 345 pages.


\subsubsection{D\textsc{iva}Services}
With D\textsc{iva}Services \cite{wursch2016sdk} Würsch et al.\ presented a fully open-source framework that allows to share and access document image analysis methods as a RESTful web service.
The idea is to allow the research community to simply provide access to newly developed methods via an unified API, independently from the used programming language, and therefore freeing the interested users from the burden of setting up and run a local instance after downloading the source code.
D\textsc{iva}Services supports various tools of different complexity, starting from smaller modules like binarization or line segmentation to more comprehensive tools like D\textsc{iva}nnotation \cite{seuret2018semi} which again can call other services themselves.

A recent publication \cite{wursch2018web} gives the latest updates concerning the execution environment (now using Docker just like OCR4all), the asynchronous execution of services, the output definition, and a planned workflow system that should allow the users to create their own workflows by specifying which modules, tools, and processes should be called in which order and with which parameters.
Furthermore, there is a focus on building an ecosystem of tools and services providing further functionality to improve the usability of the system without being part of the core framework.
This includes tools and services supporting experimentation, data and method management, programming libraries, and optimization.

While D\textsc{iva}Services is a promising approach it cannot be considered a real workflow tool and is not meant to be one, yet.
However, the available online collection of the document image analysis tool D\textsc{iva}Services Spotlight\footnote{\url{http://wuersch.pillo-srv.ch}} represents a very helpful option to perform exemplary tests of existing methods on one's projects without the need for complicated setup operations.
Unfortunately, to the best of our knowledge, not all showcased methods are  available as open source.

\subsubsection{OCR-D}
The OCR-D project\footnote{\url{http://ocr-d.de/eng}} \cite{neudecker2019ocrd} is funded by the German Research Foundation (Deutsche Forschungsgemeinschaft - DFG), initially for a period from 2015 to 2020.
Its main goal is to provide an OCR workflow for historical printings starting from the 16th century.
The workflow defines a number of modules which are executed sequentially and whose development by different German research facilities is also funded by the DFG since 2018.
Since the focus lies on mass digitization the aim is to keep the amount of manual user interaction to a minimum, ideally reducing it to zero resulting in a fully automatic workflow.
Therefore, book-specific training or any kind of manual postcorrection, be it on layout or textual level, are currently neither envisaged nor desired.
However, just as with Kraken, the high degree of modularity makes OCR-D an interesting project whose further developments should be closely followed.
This is especially true since OCR-D also relies on PageXML and therefore has publicly released several wrappers for tools like Tesseract 4 to fully integrate them into their workflow.
Since the project and therefore the developments of the submodules are still ongoing no evaluations of the overall workflow have been published, yet.

Additonally to the efforts described above, OCR-D also aims to provide a GT reference corpus for German texts printed between 1500 and 1900.
A description of the necessary formats and guidelines is given by Boenig et al.\ \cite{boenig2019gt}.

\subsection{Further OCR Related Tools}

In the following we will briefly introduce a selection of tools which are relevant to the OCR4all workflow, either because they serve as an additional preparation instance, are already integrated as submodules or represent an interesting option for future developments.

\subsubsection{Scan Tailor}
ScanTailor\footnote{\url{https://scantailor.org/}} is an interactive open-source postprocessing tool for scanned pages which offers a variety of tools and routines allowing to prepare scans for further processing.
Among others this includes:

\begin{itemize}
    \item Splitting pages that have been scanned together into two single pages.
    \item Rotating scans that are available in landscape format into an upright position.
    \item Deskewing.
    \item Removing the scan periphery and cutting out the print space.
    \item Converting the images into binary.
    \item Several smaller preprocessing techniques like despeckling and dewarping.
\end{itemize}

Most of the steps can be performed manually or by fully automatic routines whose effectiveness highly depends on the specific task and the material at hand.
For example, while the deskewing works very reliably, cutting out the print space produces severe errors on complicated layouts on a regular basis.
Nevertheless, the majority of steps can usually be performed with limited manual interaction and manageable human effort.
Unfortunately, the development of the project appears to be dormant.

\subsubsection{LAREX}
For the task of region segmentation we rely on LAREX (Layout Analysis and Region EXtraction; see \cite{reul2017larex}), which is a semi-automatic open-source tool for layout analysis on early printed books.
The primary goal of LAREX is not to fully automatically achieve a decent standardized result but to enable the users to obtain their personal 100\% result with manageable effort.
This is particularly true for the desired complexity of the segmentation, that is the degree of semantic distinction within the result.
Therefore, LAREX relies on a simple, yet effective rule-based approach which uses connected components (CCs).
Furthermore, it is very fast, easily comprehensible for basically any given user, and allows an intuitive manual correction if necessary.
To minimize the amount of required manual effort the user can define book-specific masks which represent expectations regarding the size and position of layout elements.
The results are stored using the PageXML format to support integration into existing OCR workflows.
Evaluations showed that LAREX provides an efficient and flexible way to segment pages of early printed books:
On a case study using an early printed book with complex layout it took a human processor with some prior experience using LAREX 2 hours and 18 minutes to segment the entire book including a fine-grained semantic distinction of layout elements.
For comparison, a processor with extensive experience using Aletheia \cite{clausner2011aletheia} only managed 160 page (28\% of the entire book) during the same time frame.

\subsubsection{PoCoTo}

In the context of historical OCR the interactive postcorrection tool PoCoTo\footnote{\url{https://github.com/cisocrgroup/PoCoTo}} represents the state-of-the-art.
The original PoCoTo introduced by Vobl et al.\ \cite{vobl2014pocoto} is a system developed to support the efficient interactive postcorrection of historical texts by offering several advanced features:
Suspicious tokens of the OCR text are identified by a special language technology which is aware of historical language variations represented by rewrite rules like $t\rightarrow th$ (modern spelling vs. historical spelling) and can be corrected by choosing a word from a list of generated plausible correction candidates.
The user does not have to perform this for every single word but can batch correct entire error series which for example can consist of identically misrecognized words or words that suffer from the same OCR error, for example the confusion of ``e'' and ``c''.
At any time it is possible to view the corresponding words within the scanned image.
Evaluations performed in three major European libraries covering historical German, Spanish, and Dutch showed that even without the batch correction PoCoTo already helps to efficiently correct texts.
In a first settings the users were only allowed to use the GUI which highlighted non dictionary words but could not access the batch mode.
The full mode was enabled as a second setting.
While the users on average were able to correct 3.8 errors per minute in the first setting this number almost doubled to 7.5 in the second setting.
The real potential of the software became apparent when a user took advantage of some very productive error patterns, resulting in about 500 corrections in ten minutes.

The system is under active development which resulted in several improvements on the original approach.
In \cite{fink2017profiling} Fink et al.\ added three major extensions:
First, making the system more adaptive to manual interventions of the user increased the precision with respect to identifying erroneous OCR tokens.
Second, the linguistic background resources were extended by new historical patterns which leads a more successful discrimination of historical spelling from real OCR errors.
Third, tokens that could not be interpreted by the model were added to a list of conjectured errors, resulting in a better error detection recall and precision.

A fully automated extension of PoCoTo was proposed by Englmeier et al.\ \cite{englmeier2019aipocoto}:
A-PoCoTo is more geared towards the deployment in large-scale digitization projects, can take the OCR results of multiple OCR engines into account, and uses sentence context for its decisions.
This fully automatic first step is extended by an interactive postcorrection (resulting in A-I-PoCoTo) as a second, optional step in which the user can efficiently confirm, reject, or improve decisions made by the system.

Whereas the original PoCoTo client was a stand-alone Java application it has now been rewritten as a web-based tool which together with the interactive aspect and the focus on historical texts makes it a very interesting candidate for the integration into the OCR4all workflow.

\subsubsection{Nashi}
The Nashi\footnote{\url{https://github.com/andbue/nashi}} \cite{buettner2019nashi} transcription environment was created as a platform for the digitisation of the Arabic-Latin translations corpus\footnote{\url{http://arabic-latin-corpus.philosophie.uni-wuerzburg.de}} (ALC) at the
University of Würzburg.
Its main focus was to provide a group consisting of researchers and students with the opportunity to collaboratively segment, transcribe, and comment on scans of historical and modern printed editions in Latin, Arabic, and Greek language.
Since the clear main goal is the creation of accurate citable digital editions, the web user interface for postcorrection provides the users with means to check and, if necessary, correct the OCR output for every single text line while also allowing to alter the coordinates of the line polygons.
The transcription workflow is based on PageXML and can be considerably supported by OCR processes running in the background.
The current ALC setup at the University of Würzburg relies on LAREX for segmentation, Kraken for line segmentation, and Calamari for the OCR.


%% file: chapters/03_methods.tex
\section{Methods}
\label{sec:ocr4all}

In this section we focus on the OCR4all software.
After introducing the data structure we first describe the workflow and its individual modules, including their input/output relations, in detail before we look at the encapsulating web GUI which offers various possibilities to influence the workflow by manual corrections or configurations.

\subsection{Data Structure}

Regarding our data structure, apart from different representations of page images we focus on PageXML \cite{pletschacher2010page} as the main carrier of information.
This allows for a modular integration of the main submodules of OCR4all and sets up easy to fulfill requirements regarding interfaces, ensuring a reasonably straightforward addition of new submodules or replacement of existing ones.
Additionally, defining a unified interface for all tools and modules enables the usage of a comprehensive post processing functionality.
Another positive side effect of this approach is that submodules developed by us for OCR4all can be integrated analogously into other OCR workflows that use PageXML.

PageXML requires one XML file per page which can store a wide variety of information, most importantly:

\begin{itemize}
    \item A page can comprise an arbitrary number of regions whose reading order can be specified.
    \item Among others a region can store its enclosing polygon and type.
    \item There are main types like image, text, or music. Text regions can be further classified into sub typed like running text, heading, page number, marginalia, etc.
    \item A region can contain an arbitrary number of text lines.
    \item Each line stores its enclosing polygon as well as an arbitrary number of text elements which may contain GT, various OCR outputs, normalized texts, etc.
\end{itemize}



During the setup procedure a database containing two folders, \textit{data} and \textit{models}, is mounted from the host system into the docker container.
While the first one allows to directly add new input files into the system and provides access to the final output as well as the interim results the second one enables the user to import external OCR models or to extract trained models, for example in order to share them with fellow OCR4all or Calamari users.
To add a new book the user simply has to create a new project folder within the \textit{data} folder including an \textit{input} folder containing the scans as single images or as one or several PDF files.
During the processing of a step the resulting images and PageXML files (see the individual modules below for a detailed description) are kept in a \textit{processing} folder before the final results can be generated as \textit{output}.

\subsection{OCR4all Workflow}

\begin{figure}[!htb]
    \centering
    \includegraphics[width=\linewidth]{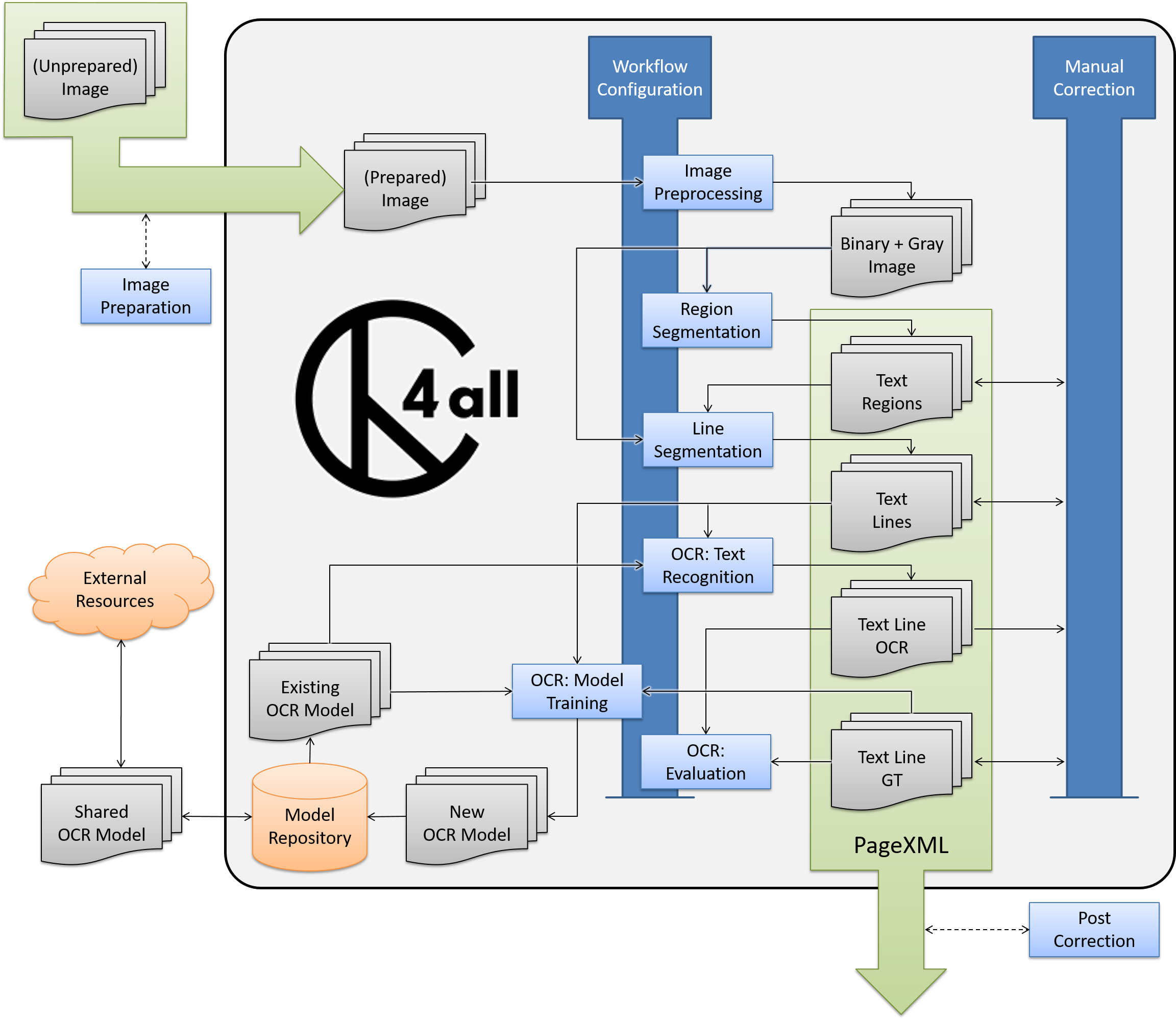}
    \caption{The main steps of the OCR4all workflow as well as the optional image preparation and post correction steps which are not part of the main tool (yet).}
    \label{fig:workflow}
\end{figure}

Figure \ref{fig:workflow} shows the steps of the workflow implemented in OCR4all.
After acquiring the scans and an optional preparation step, for example by using ScanTailor, the original images can be placed into the workspace.
Next, image preprocessing is applied to the scans before several steps, like region segmentation and extraction as well as line segmentation, produce line images required as input for character recognition or ground truth production.
The output of character recognition can either directly serve as the final result or can be corrected by the user which enables a training of more accurate book-specific models, yielding better recognition results.

In the following we discuss a typical workflow by going through the four main steps from Figure \ref{fig:schema} and discuss the corresponding modules in OCR4all as shown in Figure \ref{fig:workflow}.
Furthermore, we always state the input and output relation of each module by describing the actual data each module works on, which is often produced by combining the information stored as PageXML with the preprocessed grayscale or binary image.

\subsection{Preprocessing}
In the preprocessing main step the input images are prepared for further processing.
Before the two standard sub tasks binarization and deskewing take place an optional external preparation step can be performed.

\subsubsection{Image Preparation}
\noindent\textbf{Input}: unprepared image (image containing two scanned pages, a page rotated in an invalid orientation, ...)\\
\textbf{Output}: prepared image (single page in an upright position)

\begin{figure}[!htbp]
    \centering
    \includegraphics[width=\linewidth]{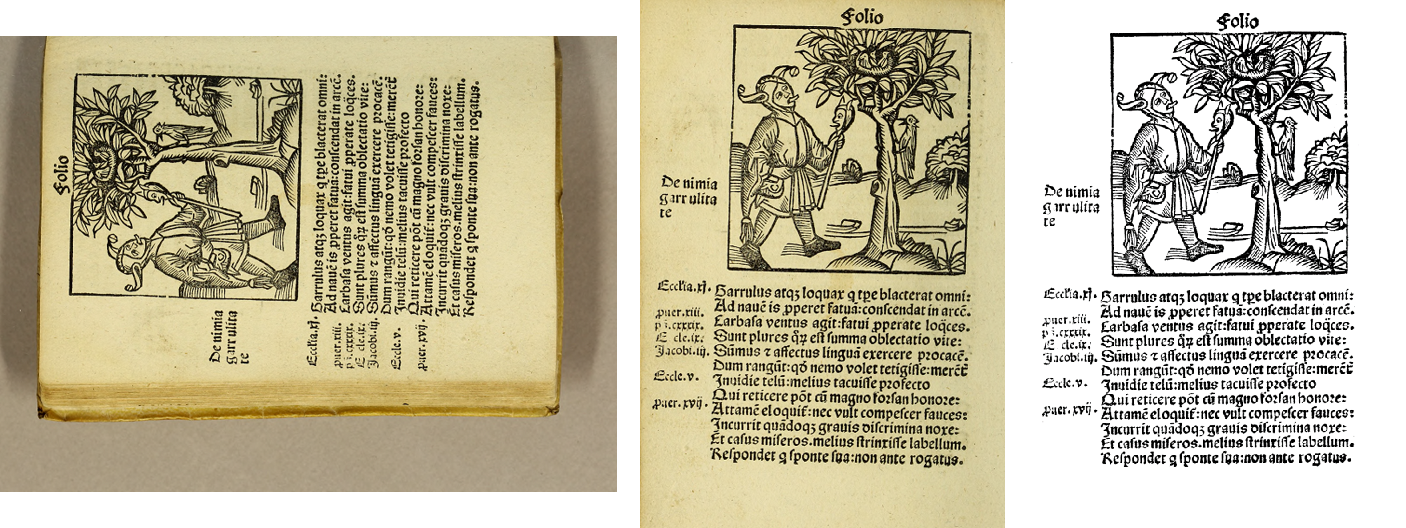}
    \caption{An example input and output of the image preparation and preprocessing steps.
    The image on the left represents an undesirable input for OCR4all while the ScanTailor output in the middle is completely sufficient.
    During the preprocessing step the skewed color image in the middle is transformed into the deskewed binary image on the right.
    }
    \label{fig:scan_prep_pp}
\end{figure}

OCR4all expects the input images to be in an upright position and already segmented into single pages which can easily be achieved by using ScanTailor.
Furthermore, it is recommended to remove excessive amounts of scan background, although this is not mandatory.
Figure \ref{fig:scan_prep_pp} shows an example of a valid and an invalid input which also represents a possible input and output of ScanTailor.
In fact, ScanTailor is not a true OCR4all submodule since it cannot be integrated due to the lack of a web-based user interface.
However, we still decided to list it as a module since this step belongs to the workflow and the input images have to be added from external sources anyway.
It is possible to deal with unprepared images like the ones described in the input completely within OCR4all but it certainly is not the recommended course of action.

During the preprocessing sub step the input image gets converted into a binary and (optionally) a normalized grayscale image.
Additionally, a deskewing operation can be performed.
See Figure \ref{fig:scan_prep_pp} for an example input and output of this step.
For both steps we use the methods implemented in the OCRopus 1 \textit{nlbin} script.
The binary/grayscale conversion is performed by applying an adaptive binarisation technique proposed by Afzal et al.\ \cite{afzal2013robust} which is able to reliably produce high quality results even when facing difficult conditions, such as heavily degraded scans with considerable brightness variations and bleed through.
While this substep is mandatory to enable and facilitate the upcoming image processing applications, the deskewing on page level is optional since its main purpose is to support the line segmentation process (see below) that operates on segmented regions which are individually deskewed beforehand anyway.
Yet, depending on the degree of skewness of the original scans, it can be very beneficial to perform a deskewing already on page level as it can considerably simplify the region segmentation.

\subsection{Segmentation}
During the segmentation main step the preprocessed images are first segmented into regions.
Then, after extracting the ones containing text the line segmentation is performed.

\subsubsection{Region Segmentation}
\noindent\textbf{Input}: preprocessed image\\
\textbf{Output}: structural information about regions (position and type) and their reading order

The general goal of this step is to identify and optionally classify regions in the scan.
There are different manifestations which considerably impact the complexity of the task and entirely depend on the material at hand, the use case, and the individual requirements of the user.
For example, when the goal is to gather and process a book for editorial purposes it is mandatory to obtain a flawless text and consequently a (close to) flawless segmentation result is advisable.
Additionally, in these cases it is often desired to perform a semantic classification of text regions already on layout level.
Therefore, a considerable amount of human effort has to be expended.
On the contrary, the most simplistic approach would be to only distinguish between text and non-text regions in order to obtain a good OCR result, for example to be used in different NLP tasks which do not require flawless texts.
Naturally, this is a considerably less costly process and can even be a trivial task, depending on the input material.
For both scenarios OCR4all offers viable solutions which will be briefly explained in the following.

\begin{figure}[!htbp]
    \centering
    \includegraphics[width=0.7\linewidth]{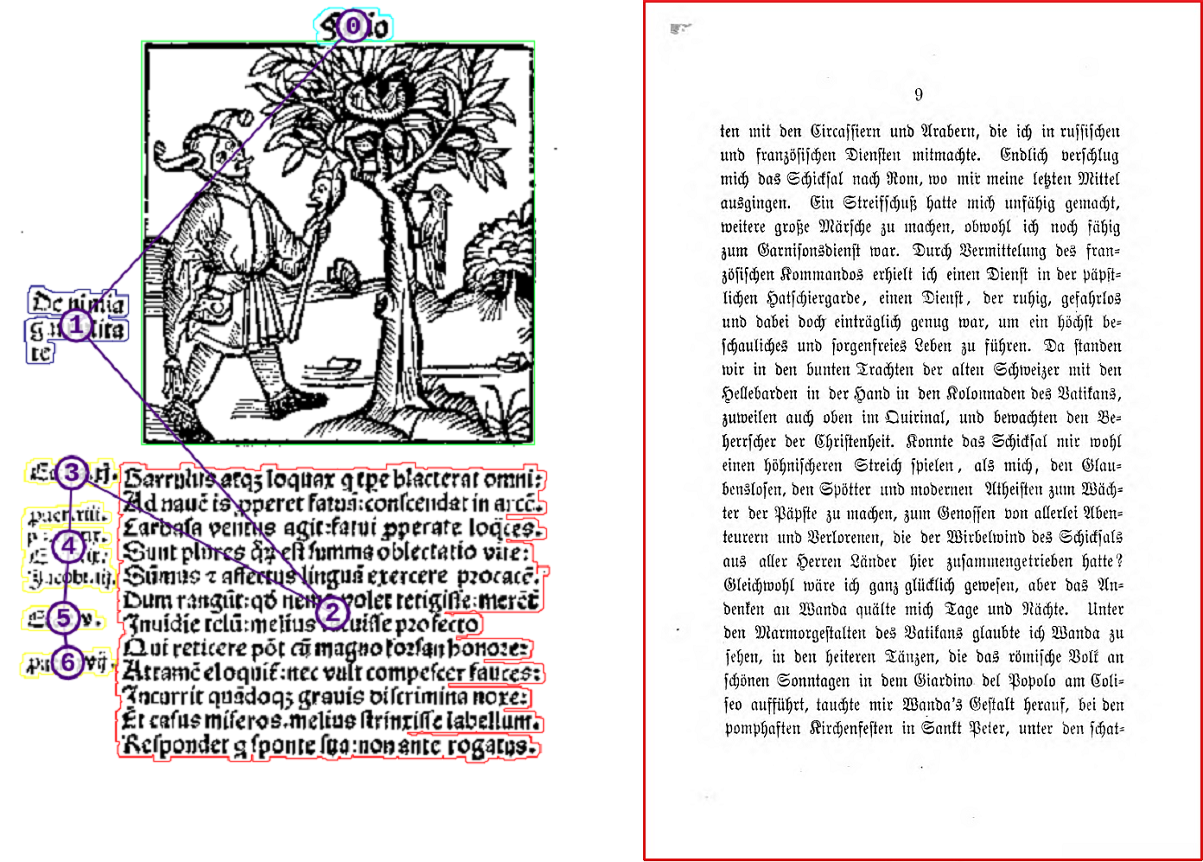}
    \caption{Left: a LAREX segmentation output consisting of an image (green), running text (red), marginalia (yellow), image caption (blue), and the page number or folio identifier (cyan), as well as the reading order. Right: output of the Dummy Segmentation for a standard 19th century novel layout.}
    \label{fig:region_seg}
\end{figure}


For complex layouts and especially when a fine grained semantic distinction is desired (see Figure \ref{fig:region_seg}, left), the already introduced tool \textbf{LAREX} represents the mean of choice.
It offers the user a variety of automatic, assisted, and fully manual tools which allow to gather a complex page layout with reasonable effort.
It is worth mentioning, that it is also possible to load existing segmentation results into LAREX and mostly use it as an editor by comfortably correcting the results, if necessary.
The drawback of LAREX is that, at least as of now, it expects the user to have a look at every page and approve each result individually.
Obviously, while checking each page is almost imperative for very complex layouts and high user expectations, it cannot be considered the preferred solution for considerably easier or even trivial layouts.
For this other end of the spectrum, OCR4all offers the so-called \textbf{Dummy Segmentation} which simply considers the entire page as a single running text segment (see Figure \ref{fig:region_seg}, right).
While admittedly this is a highly simplified approach, our experiences using the Dummy Segmentation have been very positive for several reasons.

For example, there is a rudimentary implicit text/non-text segmentation available as well as a highly performant column detection functionality.
Furthermore, because of the aforementioned capabilities of the line segmentation, the Dummy Segmentation can actually be applied to an unexpectedly wide variety of historical printings and, in fact, runs fully automatically and basically at no cost.
On the downside, this approach does not perform any kind of meaningful semantic distinction of text parts and also cannot provide an explicit image markup.

Naturally, the user can decide on a page by page basis which segmentation approach to apply.
For example, if a book starts with a quite complicated title page and a complex register, both in terms of layout, but apart from that consists of pages with a trivial one column layout, the user can easily segment the first few pages using LAREX and then switch to the dummy segmentation for the remainder of the book.
Due to the well defined interfaces all preceding and subsequent steps can be applied to all pages in the exact same manner and without any further differentiations.

\subsubsection{Line Segmentation}
\noindent\textbf{Input}: text region regions\\
\textbf{Output}: extracted text lines

\begin{figure}[!htbp]
    \centering
    \includegraphics[width=\linewidth]{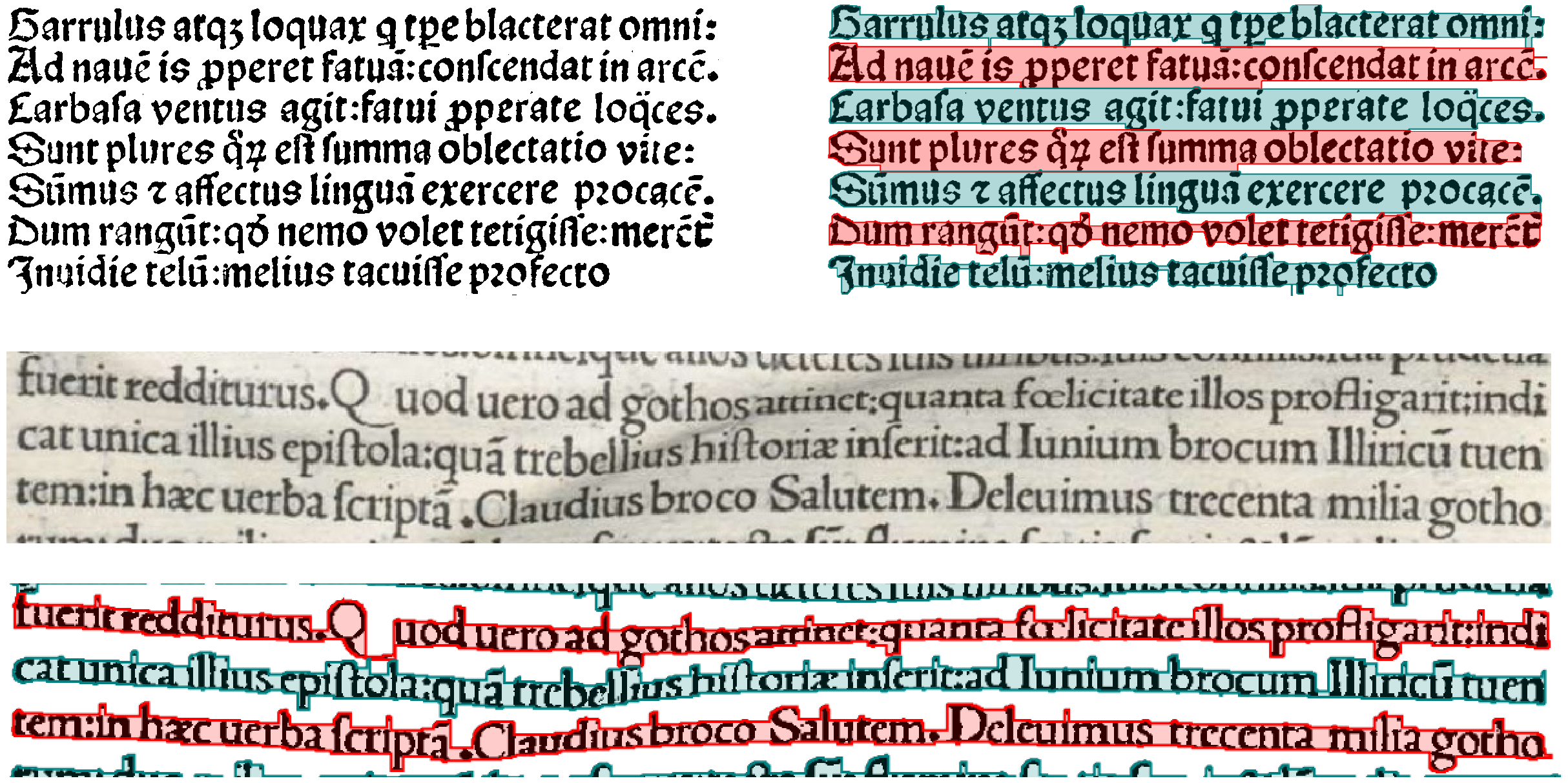}
    \caption{Two example inputs and outputs of the line segmentation step.}
    \label{fig:line_seg}
\end{figure}

The actual line segmentation operates on individual region images, if available, instead of the entire page.
Therefore, the text regions identified during the segmentation step need to be extracted from the page images which is done by a region extraction sub step. We cut out the polygons stored in the PageXML file from the corresponding binary image.
By extracting the exact polygon instead of just the bounding rectangle we ensure that even complex alignments of several regions can be processed without any overlap of other regions.
After the extraction, the region images are separately deskewed by applying the OCRopus 1 \textit{nlbin} script.
Processing the regions one by one can lead to considerably better results than the standard deskewing on page level.
Especially pages from very old printings frequently contain areas/regions which are skewed independently from each other, either because of inaccuracies during the printing process or due to physical degradation leading to deformed pages.
Clearly, a globally optimal skewing angle on page level cannot deal with these situations.

The line segmentation is performed by applying an adapted version of the Kraken line segmentation script to each extracted region individually.

The output produced by the applied algorithm is considerably more complex than just the bounding rectangle of a text line:
After eliminating noise and other unwanted elements, the connected components are assigned to their respective text lines, mostly as a whole but can also be split, if necessary.
Finally, the connected components or their respective parts are connected to a tight-fitting polygon in order to produce an optimal line segmentation result, exclusively containing the desired letters.
To achieve this we extended the \textit{Nashi}
line segmentation wrapper.
Analogously to the region segmentation step the usage of exact polygons even allows to segment lines whose bounding rectangles considerably overlap.
An extreme example of the segmentation capabilities is shown on the bottom of Figure \ref{fig:line_seg}. The following OCR step had no difficulty with the recognition of each separate line.

\subsection{OCR}
After obtaining the text lines the OCR main step can be performed including the character recognition either by applying available mixed models or the results of a book-specific training.
Furthermore, an error analysis of produced outputs is provided which just like the training requires GT that can be produced by manual correction which we will discuss later on.

\subsubsection{Character Recognition}
\noindent\textbf{Input}: text line images and one or several OCR models\\
\textbf{Output}: textual OCR output on line level

After segmenting the pages into lines it is now possible to perform OCR on the results.
As of now, Calamari is the only OCR engine which is integrated into OCR4all by default.
However, due to the well defined interfaces additional engines can be added and operated with manageable effort.

For the application in OCR4all we make use of Calamari's PageXML interface which  cuts out the text lines from the corresponding binary or grayscale images according to their coordinates and passes them into the recognizer.
In general, the recognition module allows to either apply self trained book specific models, which we will address in the next section, or to resort to so-called mixed models.
These models have been trained on a wide variety of books and typesets and, depending on the material used, can usually provide at least a valid starting point to start off the manual GT production or even already provide a satisfactory final result.
OCR4all comes with four single standard models\footnote{\url{https://github.com/Calamari-OCR/ocr4all_models}} which are automatically incorporated and made available when building the Docker image: antiqua\_modern, antiqua\_historical, fraktur\_19th\_century, and fraktur\_historical.
Since voting ensembles have proven to be very effective, we additionally provide a full set of model ensembles\footnote{\url{https://github.com/Calamari-OCR/calamari_models}} consisting of five models for each of the four single model areas mentioned above, which can be downloaded and directly added into OCR4all.




Calamari supports the utilization of an arbitrary number of models.
If only one model is applied, its output is directly considered as the final result and is consequently added to the corresponding line element in the PageXML file.
The application of several models automatically triggers the confidence voting procedure where the final result is calculated from the single outputs of all the voters.
Apart from the standard textual output it is also possible to enable an additional, extended output that includes information like the intrinsic OCR confidence values calculated by Calamari or the pixel positions of the detected characters within the line.
Parts of this data could also be stored by using PageXML but since we want to keep as much information as possible, including detailed and comprehensive lists with character alternatives and their respective confidences, an additional storage format (JSON) is required.

After transcribing lines or by correcting an existing OCR output, which we will cover in detail in the section on manual corrections below, the model training can take place.

\subsubsection{Model Training}
\noindent\textbf{Input}: line images with corresponding GT, optionally already existing models to build upon    \\
\textbf{Output}: one or several OCR models

The model training which allows to train book-specific Calamari models is not only one of the most central modules of the entire workflow but also probably the most complex and challenging one when it comes to enabling non-technical users to utilize all available features.
The training algorithm possesses a variety of hyper-parameters which can impact the procedure considerably and therefore has to be treated with great care.
To begin with, one crucial aspect when dealing with the training of neural networks there is the omnipresent challenge of determining when to stop the training process and choosing the best model, which is the one with the optimal weight configuration.
To avoid overfitting we follow the established approach of setting aside a small chunk of the training material as a validation set which is realized by passing appropriate parameter settings into Calamari.
Then, after a certain number of training iterations the current model is applied to and evaluated on this validation set.
The model which performed best is denoted as the best model and stored for further processing.
Regarding the termination of the training we make use of the so called early stopping provided by Calamari that basically observes the training progress and aborts as soon as no significant improvement can be expected anymore.
This procedure requires several parameters which determine the stopping criterion and how frequently to check the current model.
To guard the users from having to deal with the underlying theoretical concept but still ensure sensible parameters, we developed a routine that derives fitting values for all required settings from the available amount of training data and that was incorporated into Calamari.
Naturally, experienced users are free to adjust the parameters directly at will.

Our main goal for the training module was to provide a non-technical user with the ability to comfortably make use of all the available accuracy improving techniques:

\begin{itemize}
    \item Cross fold training produces a voting ensemble consisting of several individual models whose output is then combined during the prediction as explained in the recognition section.
    In order to significantly improve the recognition accuracy, the models have to be both strong individually and diverse enough with different strengths and weaknesses.
    Therefore, the training and validation data has to be arranged in a certain way into different folds which is carried out automatically by Calamari.
    All the user has to do is to determine the number of desired voters.
    \item To speed up the training and to further optimize the resulting models the Calamari training can build from already existing models instead of starting from scratch.
    Naturally, this can and should be combined with training of several models to form a voting ensemble.
    Consequently, the user can choose between three training approaches: training all models from scratch, training all models starting from the same existing models, or freely assigning arbitrary models to each fold.
    All this can be done by comfortably selecting the desired models from a list displaying all available options.
    \item Data augmentation is a powerful option to make the most of small amounts of training material.
    The basic idea is to generate additional training examples by synthetically altering existing line images, for example by applying scaling or warping operations, which still fit the original textual GT.
    To activate this in OCR4all the user simply has to determine the extent of desired augmentations.
    Since nothing compares to real data the default training allows for a two step approach in which the models resulting from training on both real and augmented data are further refined by training exclusively on real data.
\end{itemize}

\paragraph{Iterative Training Approach}
To keep the manual effort to a minimum, we introduced an iterative training approach which is fully supported by OCR4all.
The general idea is to minimize the required human workload by increasing the computational load.
Correcting an existing OCR output with a (very) good recognition accuracy is (considerably) faster than transcribing from scratch or correcting a more erroneous result.
Consequently, we aim to quickly get to a reasonable recognition accuracy which allows for an efficient GT production process.
Therefore, we integrated an iterative training approach whose procedure is listed in the following:

\begin{enumerate}
	\item Transcribe a small number of lines from scratch or correct the output of a suitable mixed model, if available.
	\item Train a book specific model/voting ensemble using all available GT that has been transcribed up to this point, including earlier iterations.
	\item Apply the model/voting ensemble to further lines.
	\item Correct the output.
	\item Repeat steps 2-4.
\end{enumerate}

For example, let's assume that around 200 lines of GT are needed to reach a satisfactory OCR accuracy (e.g., CER of 2\% or less) for a given task and on a given book.
Of course, the exact number of lines required is never known, which is another reason to not simply transcribe 200 lines from scratch before training the first model.
Instead, we start out by recognizing a small number of lines, for example two pages comprising 60 lines, with a somewhat suitable available mixed model, resulting in a more or less helpful OCR output with a CER of, let's say 8\%.
While the recognition quality is not perfect, correcting this output is still faster than transcribing from scratch and it only needs to be done for 60 lines anyways.
Next, a first book-specific model is trained by using the produced GT.
The resulting model is then applied to four more page, resulting in a considerably lower CER of 3.5\%.
In fact, correcting this output can be done much more efficiently than before since the error rate was reduced dramatically by more than 50\%.
The resulting four pages of GT are now added to the GT pool (now containing close to 200 lines) and used for another training process, resulting in a strong model yielding a CER of just below 2\% on previously unseen pages.
Since this final model fulfills the requirements set for this example, the iterative training process stops here.
Obviously, the described steps can easily be repeated until a higher desired accuracy is reached or even until the entire book has been recognized and corrected.

Since the iterative training approach, especially when combined with cross fold training, can quickly produce plenty of OCR models, a certain amount of bookkeeping is required to stay on top of things.
Therefore, OCR4all provides an intuitive automatic naming convention for the trained models.

\subsubsection{Error Analysis}
\noindent\textbf{Input}: line-based OCR predictions and the corresponding GT    \\
\textbf{Output}: CER and confusion statistics

To enable an objective assessment of the recognition quality achieved by the models at hand, we incorporated the Calamari evaluation script into OCR4all.
For a given selection of pages it compares the OCR results to the corresponding GT and calculates the CER using the Levenshtein distance.
Additionally, a confusion table (see Figure \ref{tab:conf-table} for an example) displaying the most common OCR errors and their frequency of occurrence is provided.

\begin{table}[!htbp]
\centering
\caption{Example confusion table showing the desired output (\textit{GT}), the actual output (\textit{OCR}), the absolute number of occurrences (\textit{CNT}), and the corresponding percentage with respect to all errors (\textit{PERC}).
Given are the five most frequent errors including substitutions (4), deletions (2,5), and insertions (1,3).}
\label{tab:conf-table}
\begin{tabular}{c|cc|cc}

\toprule

\textbf{ID} & \textbf{GT} & \textbf{OCR} & \textbf{CNT} & \textbf{PERC}\\

\midrule

1 &   & ␣ & 16 & 6.69 \\
2 & ␣ &   & 10 & 4.18 \\
3 &   & i & 10 & 4.18 \\ 
4 & e & c &  3 & 1.26 \\
5 & l &   &  2 & 0.84 \\

\bottomrule
\end{tabular}
\end{table}

\subsection{Result Generation}
\noindent\textbf{Input}: GT and OCR results\\
\textbf{Output}: final output as text files

For the average OCR4all user PageXML most likely does not represent the desired output format that is needed for further processing with other tools.
Consequently, we also offer a simple textual output where the line-based OCR results are concatenated in reading order and stored as a text file in two variants, one for each individual page and one for the entire book.
If there is GT available for a line its OCR result is replaced by the corrected text in the final output.

Naturally, the conversion into raw text leads to the loss of all acquired additional information obtained during the complete workflow like semantic labels and coordinates of segments and lines.
To preserve this data it is of course also possible to keep the PageXML files containing all information acquired during the workflow.
Additional output formats, for example TEI\footnote{\url{https://tei-c.org}}, can easily be added.

\subsection{Web GUI}

One of the main goals of OCR4all is to allow anyone to perform OCR on their own on a wide variety of historical printings and obtain high quality results with reasonable time expenditure.
Therefore, the tool has to be easily comprehensible even for users with no technical background.
In fact, this includes the ability to comfortably control the entire process via a GUI.
By making all submodules accessible from a clearly structured and unified interface it ensures that the user for the most part has to learn only a single system.
Furthermore, OCR4all equips the user with powerful tools to perform manual corrections on the produced output after most steps of the workflow and allows for a precise configuration not only of the workflow in total but also of the sub modules.
Before we discuss these options we briefly introduce some general aspects about the tool's architecture.

\subsubsection{General Software Design}
We chose to implement the workflow as a server application accessible by a web app because this allows a deployment as a true web app (using a web browser as a local client interacting with a remote server) as well as using OCR4all completely locally (both browser and server are locally installed).
Because currently neither user administration nor resource management have been implemented, in the following we only consider the local option. 

To ensure an easy installation procedure and keep the problems caused by dependency requirements of the sub modules to a minimum, we encapsulated everything in a Docker image.
Furthermore, the incorporation of Docker effectively assures platform independency as it can be installed and run on basically all modern operating systems including Windows, Mac, and Linux.

\subsubsection{Manual Corrections}
\noindent\textbf{Input}: images and their corresponding PageXML files    \\
\textbf{Output}: corrected PageXML files\\

\begin{figure}[!htb]
    \centering
    \includegraphics[width=0.8\linewidth]{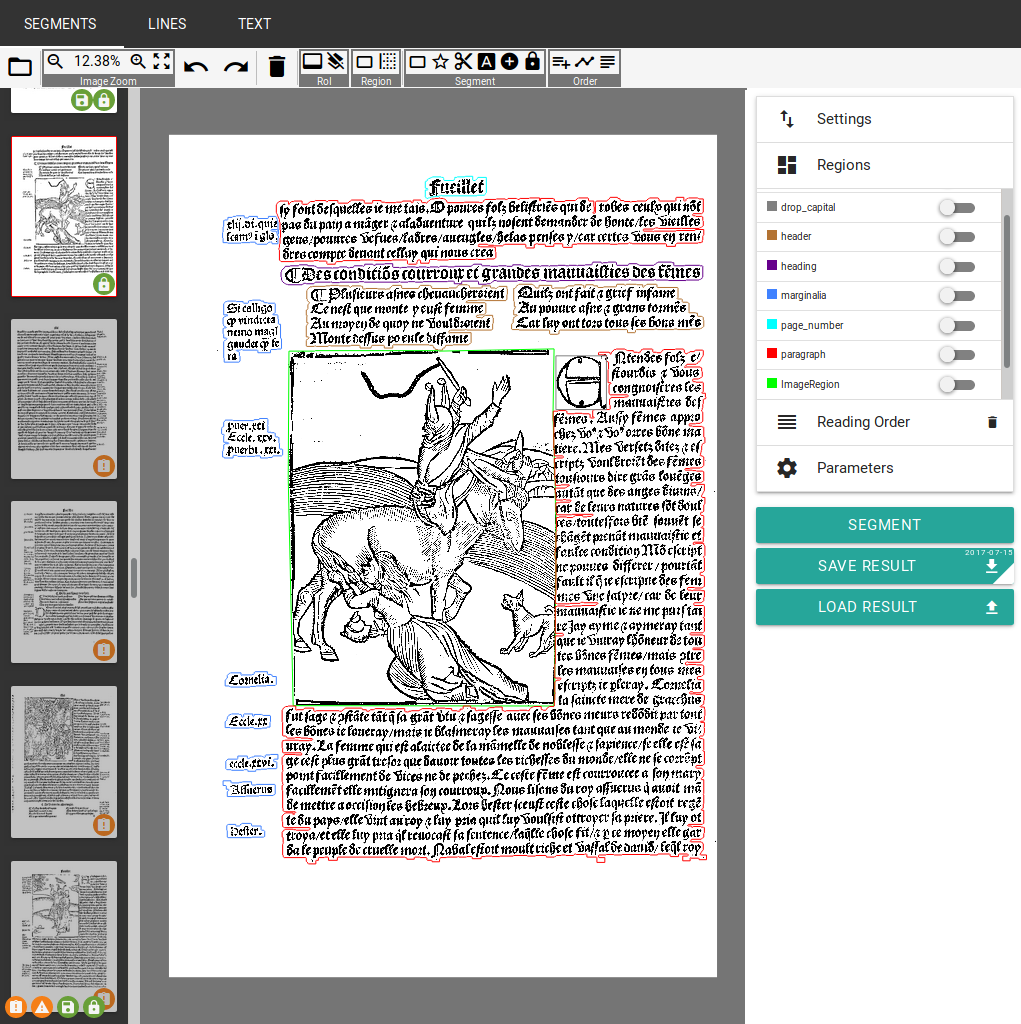}
    \caption{Compressed Overview over the LAREX correction GUI with the actual page and its current data in the middle, the page selector on the left, the three correction tabs to switch between the segments, lines, and text correction functionality as well as several tools on the top, and the settings on the right.}
    \label{fig:larex_overview}
\end{figure}

As emphasized during the introduction a fully automated workflow is often not reasonable or at least cannot be expected to yield sufficient (depending on the use case) or even perfect results especially when dealing with early printings.
Consequently, a potent, flexible, comprehensible, and easy to use option for manual correction is a must-have for every OCR workflow tool which relies on user intervention.
In OCR4all this core task is covered by LAREX whose functionality has been considerably extended since its original release \cite{reul2017larex} as a region segmentation tool and will be explained in the following (see Figure \ref{fig:larex_overview} for an an overview of the LAREX correction tool).

LAREX works directly on PageXML files and the corresponding images.
After loading a page the information is displayed using additional layers over the image in three different views which are all interconnected with each other.

\begin{figure}[!htbp]
    \centering
    \includegraphics[width=\linewidth]{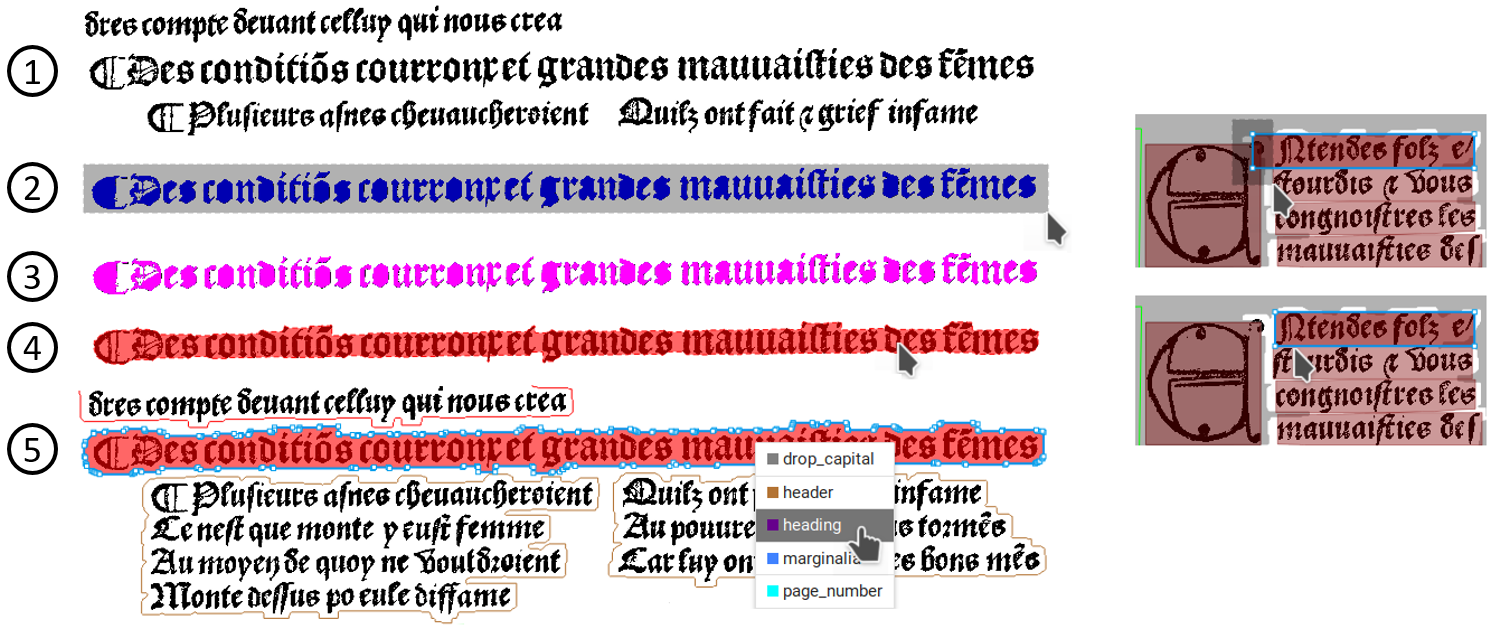}
    \caption{Example applications of the labelling and segmentation functionality on layout level in LAREX.
    Left: Semi-automatic segmentation of a complex layout (from top to bottom): (1) excerpt from the starting situation showing four lines belonging to four different segments with three different semantic types (see the left of Figure \ref{fig:larex_text} for the bigger picture).
    (2) Manual selection of the CCs belonging to the subheading.
    (3) Marked depiction of the selected CCs which serve as input for the iterative smearing algorithm.
    (4) Output region of the smearing algorithm.
    (5) Adjustment of the region type; adjacent the remaining regions which have been automatically detected (remaining lines).
    Right: correction of a line polygon which wrongfully includes part of a swash capital by moving two points.
    }
    \label{fig:larex_seg}
\end{figure}

\paragraph{Regions}
LAREX offers a wide variety of tools and procedures to create new and edit existing regions.
Regions identified during the earlier region segmentation step can be deleted and their type or sub type can be changed.
If a region has not been captured the user can correct this by either manually drawing a rectangle or a polygon or by selecting the connected components belonging to the region and then activate an iterative smearing algorithm to automatically create the region outline.
Additionally, the reading order can be freely adjusted by dragging and dropping regions in an additional view.
Furthermore, it is possible to perform sophisticated polygon manipulation operations including deleting, adding, and moving points.
Each operation can always be performed on an arbitrary number of points at once.
Combined with the progressive zooming functionality this even allows for an pixel perfect segmentation, if desired.
Figure \ref{fig:larex_seg} shows some example functionality.

\paragraph{Lines}
The second view focuses on text lines.
From an editing point of view lines are treated exactly like regions and therefore allow the same comprehensive set of operations with minor adaptions, for example that newly created lines are assigned to the active regions and not to the page and the reading order functionality is available relative to a selected region.

\begin{figure}[!htbp]
    \centering
    \includegraphics[width=\linewidth]{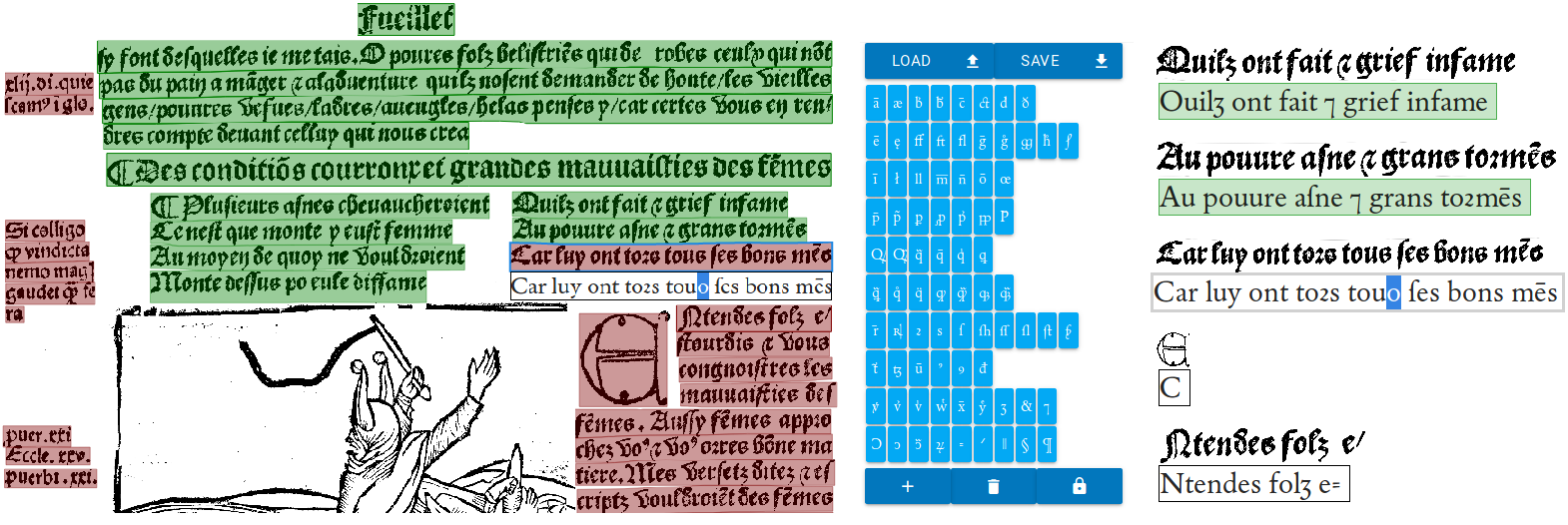}
    \caption{Two text correction views in LAREX: page based view with virtual keyboard where a selection of lines can be corrected (left and center) and the corresponding line based view (right).}
    \label{fig:larex_text}
\end{figure}

\paragraph{Text}
The text view (see Figure \ref{fig:larex_text}) is divided into two further sub views.
In the first one the page image is still presented to the user with all text lines color-coded indicating the availability of corresponding GT.
When selecting a text line an input field is displayed directly below the line, showing the corresponding OCR or GT if available.
Like the rest of the visualization the displayed textline is zoomable and can be moved horizontally or resized separately to obtain a perfect alignment with the line image since this cannot always be achieved automatically due to the narrow and often irregular typesetting. 
The user can then produce or edit GT by simply typing into the input field or by selecting characters from a customizable virtual keyboard which allows to define a set of non-standard characters, for example ligatures, which cannot be found on regular keyboards but can easily be inserted this way by a single mouse click.
The content and structure of the virtual keyboard can be customized directly in the web GUI and existing setups can be exported and imported.
Keyboard shortcuts allow to temporarily fade out the input field so the users can get contextual information from the subsequent line if desired and they may quickly cycle through the lines in reading order.

While this view is a suitable solution for users that aim for a perfect text and consequently have to take a thorough look at each line anyway, it is not optimal for use cases where users just want to quickly scan the pages and lines for obvious mistakes.
For this use case we introduced a second sub view which optimizes the correction process by providing a synoptic view where an editable text field is placed directly under each line image.
If there is already GT available for a line the corresponding text field is colored in green and the GT text is displayed.
Analogously, if there is no GT but an OCR result available, it is shown in the text field.
Otherwise, the transcription has to be performed from scratch.
When a line is selected by clicking into the corresponding text field it is marked as active and can again comfortably be edited by typing regularly or via the virtual keyboard.
When a line gets deselected, for example by activating the next line, its current content is automatically saved as GT.

Since all pairs are ordered one below the other in reading order it is possible to display the line image and the corresponding transcription for all lines at once allowing the user to get a quick overview.
Despite the completely different arrangement of the lines the interconnection with the other views still exists.
So if the users noticed a line which suffers from a serious segmentation fault they can simply switch to the line-based view, quickly identify the line since it is still marked as active, perform the necessary correction, and switch back to the text correction view to continue from where they left off.

\paragraph{Practical Integration into the Workflow}
In theory manual correction phases can of course be introduced at basically any step during the workflow.
While it is clear that ensuring optimal results after each processing step minimizes the chance for and the effect of consequential errors, a comprehensive manual inspection and correction after each step is neither required nor sensible.
For example, checking and correcting thousands of text lines one by one after the line segmentation step may well increase the achievable OCR accuracy during the subsequent recognition step.
However, a much more efficient solution is to subject a few representative pages to a quick visual check and if any systematic errors are recognized, one may use the comprehensive set of parameters to optimize the output on a global level.
In our experience with the currently available setup manual correction should only be applied directly after the region segmentation, which takes place in LAREX anyway if not performed fully automatically, or at the very end when all information including line coordinates and OCR results is available.

\subsubsection{Configurations}

To be able to deal with the wide variety of printings and the distinct challenges proposed by them, as well as to satisfy the individual needs of each user, OCR4all offers plenty of ways to influence not only the workflow in itself but also the parameters of the single sub modules.
All configurations are entirely accessible from the web GUI in an intuitive way and do not require any kind of knowledge regarding the usage of the command line.

\paragraph{Process Flow}
Most of the steps and modules introduced above give the user the chance to manually check the results and apply corrections, if necessary.
However, there are certainly plenty of use cases where this is neither desired nor necessary, for example when dealing with relatively simple layouts.
In order to enable the user to minimize the degree of manual intervention, we introduced the process flow functionality which allows to configure and execute several modules at once.
For example, when dealing with a typical 19\textsuperscript{th} century Fraktur novel as shown on the right in Figure \ref{fig:region_seg}, it may well be sufficient to fully automatically run the preprocessing, region segmentation (dummy), region extraction, line segmentation, and recognition (standard 19th century Fraktur model) steps, to obtain a high quality OCR result.
Analogously, when a segmentation using LAREX is needed, which requires user intervention, the subsequent steps can still be run at once and fully automatically.

\paragraph{Parameters}
The open-source OCR tools we utilize in our workflow often allow influencing the results by passing parameters, usually via command line options.
Optimizing the usage of these tools represented one of the main challenges during the implementation of OCR4all.
On the one hand, it is imperative not to overwhelm inexperienced users by confronting them with a plethora of confusing options, settings, and parameters.
On the other hand, it is equally vital to provide more experienced users to adapt selected settings in order to optimize the results.
As a solution, our web GUI provides interfaces to set almost all parameters of the OCRopus 1 and Calamari submodules.
Furthermore, we carefully split the available options for each submodule into general and advanced settings.
Apart from the number of threads used for execution which are by default set to the available maximum, the general settings usually only contain one or two parameters whose default settings normally completely suffice for the average user.
The advanced settings comprise all remaining parameters and allow experienced users to maintain full control.

%% file: chapters/04_evaluations.tex
\section{Evaluations}
\label{sec:evaluations}

To evaluate the effectiveness and usability of OCR4all we performed several experiments on various books using different evaluation settings which we will discuss in the following.
After introducing the data we focus on evaluating the main area of application of OCR4all, namely the precise text recognition of early printed books.
Then, we take a closer look at the effects of the iterative training approach.
Afterwards, we experiment with a reduced degree of manual intervention by the user by first evaluating a less costly but also less precise segmentation approach and then evaluate a fully automatic process on newer works.

The main goals of our experiments are to evaluate the

\begin{itemize}
    \item manual effort required to capture a book using a precise segmentation and aiming for a very low error rate (<1\% CER) dependent on the complexity of the material and the experience of the user
    \item speed up when incorporating the iterative training approach
    \item potential speed up when considerably lowering the requirements regarding segmentation, especially considering the fine-grained semantic distinction of layout elements
    \item performance of OCR4all when applied to newer works with simpler layouts
\end{itemize}

For reasons of clarity we will provide a table for each experiment and only briefly sum up the main results in this section while shifting the in-detail discussion to section \ref{sec:discussion}.

\subsection{Data}

In this section we briefly introduce the books we used for our experiments comprising a variety of early printed books and 19th century Fraktur novels.

\subsubsection{Early Printed Books}

Due to the focus of OCR4all on early printed books a large portion of our evaluation corpus consists of books printed before 1600 which are listed in Table \ref{tab:early_printed_books} and can be further subdivided into three groups.

\begin{table}[!htb]
\centering
\caption{Books of the early modern age used for our experiments including their \textit{Full Title} and the \textit{Languages} used within them.
The \textit{Identifier} encodes the group (\textbf{C}amerarius, \textbf{N}arrenschiff, \textbf{P}ractical course) and the year of publication.
}
\label{tab:early_printed_books}
\begin{tabular}{lll}
\toprule
\textbf{Identifier} & \textbf{Full Title} & \textbf{Languages} \\ 

\midrule

N1494 & Das Narrenschiff & German \\ 
N1498 & La nef des folz du monde & French \\ 
N1499 & La grant nef des folz du monde & French \\ 
N1506 & Nauis stultifera & Latin \\ 
N1549 & Der Narren Spiegel & German \\ 

\midrule

C1532a & Astrologica & Latin, Greek  \\ 
C1532b & Ioachimi Camerarii Norica sive de ostentis libri duo & Latin, Greek \\ 
C1533 & De theriacis et mithrateis commentariolus & Latin, Greek \\ 
C1535 & Erratum & Latin, Greek \\ 
C1541 & Elementa rhetoricae & Latin, Greek \\ 
C1552 & Historia synodi nicenae & Latin, Greek \\ 
C1554 & Versus senarii de analogiis & Latin, Greek \\ 
C1557 & \makecell[l]{Libellus alter, epistolas complectens Eobani et aliorum\\quorundam doctissimorum virorum} & Latin, Greek \\ 
C1558 & De eorum qui cometae appellantur & Latin, Greek \\ 
C1561 & Tertius libellus epistolarum H. Eobani Hessi & Latin, Greek \\ 
C1563 & Dialogus de vita decente aetatem puerilem & Latin, Greek \\ 
C1566a & De Philippi Melanchthonis ortu, totius vitae curriculo et morte & Latin, Greek \\ 
C1566b & \makecell[l]{Historiae Iesu Christi Filii Dei Nati In Terra Matre\\ Sanctiss. sempervirgine Maria summatim relata expositio} & Latin, Greek \\ 
C1568 & Libellus novus epistolas et alia quaedam monumenta doctorum & Latin, Greek \\ 
C1583 & Epistolarum familiarum libri VI & Latin, Greek \\
C1594 & \makecell[l]{Decuriae XXI symmikton problematon\\seu variarum et diversarum quaestionum de natura, moribus, sermone} & Latin, Greek \\ 
C1598 & De rebus turcicis commentarii & Latin, Greek \\ 

\midrule

P1474 & Das abenteürlich buch beweyset vns von einer frawen genandt Melusina & German \\ 
P1484 & Histori von herren Tristrant & German  \\ 
P1509 & Fortunatus Eyne hystorye & German \\ 

\bottomrule
\end{tabular}
\end{table}

The first group consists of editions of the \textit{Narrenschiff} (ship of fools), the second most popular book after the bible in the early modern period, and was digitized as part of an effort to support the \textit{Narragonien digital} project\footnote{\url{http://kallimachos.de/kallimachos/index.php/Narragonien}} at the University of Würzburg.
Despite their similar content these books are very different from an OCR point of view since their layout varies considerably and they were printed in different print shops using different typefaces and languages (Latin, German, French and Dutch).
In the second group we deal with printings related to the influential early modern universal scholar Joachim Camerarius the Elder whose numerous works have been identified and collected during the \textit{Opera Camerarii} project\footnote{\url{http://wp.camerarius.de/}} at the University of Würzburg.
These works, which are now intended to be captured by OCR, are mostly written in Latin but frequently contain embedded parts of Greek, mostly scientific technical terms regarding the treated topics like astrology, medicine, and many more.
A special feature of these books is that they often contain Greek sections directly within the Latin text.
We focus on the OCR of the Latin parts and just ensure to mark Greek text for later processing.
While this might seem counter-intuitive at first, OCR engines have been shown to be able to learn abstract representations of different scripts \cite{ul2015sequence} or even different but very similar fonts \cite{reul2019automatic}.

\begin{figure}[!htb]
    \centering
    \includegraphics[width=\linewidth]{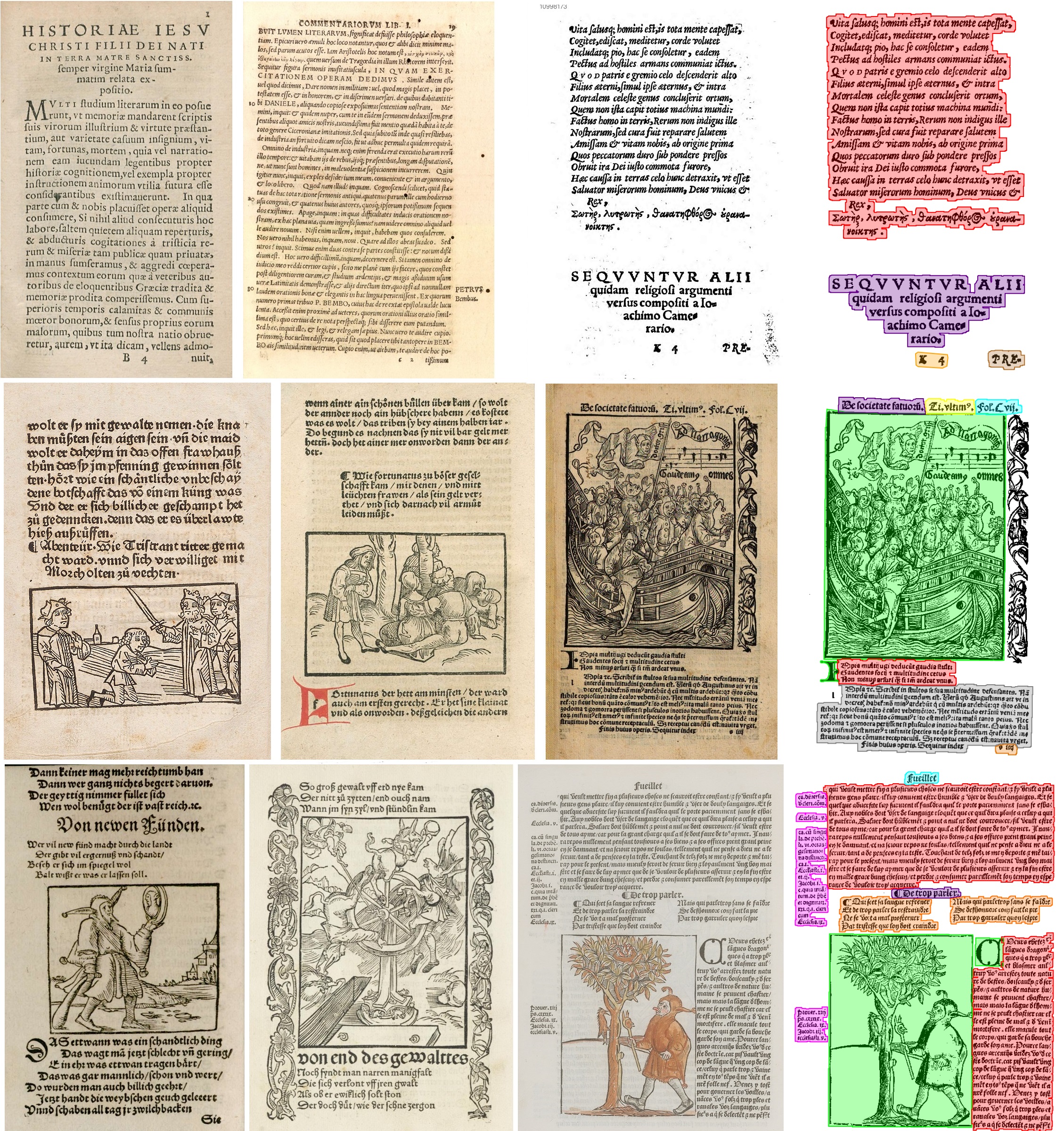}
    \caption{Example images of early printed books. Top (from left to right): C1566, C1541, C1563, C1563 segmented. Middle: P1484, P1509, N1506, N1506 segmented. Bottom: N1549, N1494, N1499, N1499 segmented.}
    \label{fig:epb-images}
\end{figure}

Finally, the third group consists of various early modern printings.
Figure \ref{fig:epb-images} shows representative example images of some of the used books as well as some desired segmentations.
For reasons of clarity we refrained from depicting the reading order.


\subsubsection{19th Century German Novels printed in Fraktur}
The second part of our evaluation corpus consists of 19th century German novels (with one exception from the late 18\textsuperscript{th} century) which are currently collected and OCRed by the Chair for Literary Computing and German Literary History of the University of Würzburg.

\begin{figure}[!htbp]
    \centering
    \includegraphics[width=\linewidth]{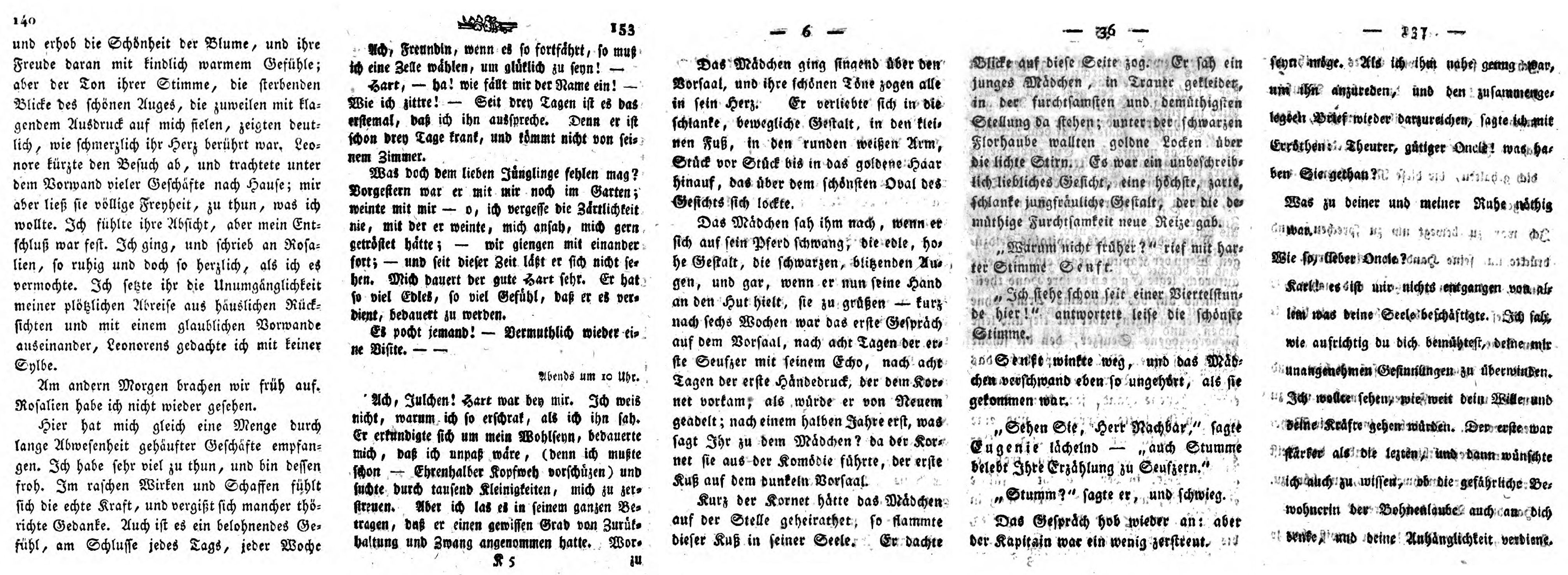}
    \caption{Example images of the german novel corpus.
    From left to right: F1870, F1781, F1818 (page in decent condition), F1818 (page in bad condition), F1803.}
    \label{fig:fraktur_imgs}
\end{figure}

Most of the books were scanned in 300 dpi and were provided by the Bayerische Staatsbibliothek\footnote{\url{https://www.bsb-muenchen.de/}}.
The overall quality of the material varies considerably as shown in Figure \ref{fig:fraktur_imgs}.
This task requires a completely different OCR approach for various reasons:
The resulting corpus is intended to be used for experiments with quantitative approaches which are usually quite robust with respect to OCR errors.
Furthermore, from an OCR point of view, the material is considerably less complex compared to the early printed books we discussed before, due to its rather trivial layout, more standardized typography, and its superior state of preservation.
The corpus is very extensive, currently comprising around 1,800 novels, and aiming to OCR all novels of this period (probably 10,000 to 15,000).

These aspects make it neither necessary nor feasible to invest an extensive amount of manual work.
A highly automated workflow is intended instead.

\begin{table}[!htbp]
\centering
\caption{German Fraktur novels used for our experiments given by their \textit{Author}s and their \textit{Title}s.
The \textit{Identifier} encodes the group (\textbf{F}raktur) and the year of publication}.
\label{tab:fraktur-novels}
\begin{tabular}{cll}

\toprule

\textbf{Identifier} & \textbf{Author} & \textbf{Title}\\

\midrule

F1781  & Friedel, Johann                    & Eleonore                      \\
F1803  & von La Roche, Sophie               & Liebe-Hütten                  \\
F1810  & Fouqué, Friedrich de la Motte      & Der Held des Nordens          \\
F1818  & Lafontaine, August Heinrich Julius & Reinhold                      \\
F1826  & Pichler, Caroline                  & Frauenwürde                   \\
F1848  & Hahn-Hahn, Ida                     & Levin                         \\
F1851  & Müller, Otto                       & Georg Volker                  \\
F1865  & Hiltl, Georg                       & Gefahrvolle Wege              \\
F1869  & von Hillern, Wilhelmine                & Der Arzt der Seele            \\
F1870  & Hiltl, Georg                       & Die Bank des Verderbens       \\

\bottomrule
\end{tabular}
\end{table}

\subsection{Precise Segmentation and Trained OCR of Early Printed Books}

In this first evaluation we will examine the performance of OCR4all on the task which represents its main area of focus: the OCR of early printed books with the aspiration to obtain a (close to) perfect result, both regarding segmentation and OCR, even if this means a substantial amount of manual work for the user.

\subsubsection{General Processing Approach}
Each book is always processed by a single user with the exception of 1494 which was cut in half and assigned to two users for independent processing.
Clear guidelines for the segmentation of any book had been specified beforehand.
The most important ones can be seen in the following:

\begin{itemize}
    \item The entire book is segmented by the user and the required time is recorded.
    \item A fine-grained semantic classification of layout elements  level is required, including the distinction of images, running text, headings, page numbers, marginalia, signature marks, catchwords, and swash capitals.
    \item After segmenting down to line level the GT production and iterative training approach starts. For evaluation purposes only the transcription of whole pages was viable.
    \item To get notable improvements during the iterative training approach the amount of added GT should rise considerably during each step.
    The suggested approach was to start out with two to four pages, then add three to five pages during the next iteration and so on.
    \item Ideally, representative pages with respect to their state of preservation, print quality, and used fonts should be selected for training and evaluation.
    Especially for the Camerarius project the font aspect is particularly important, since many books comprise large sections printed in italics while the bulk of the text is printed in an upright font.
    \item Since most of the books are meant to be transcribed in full later on in their respective projects, a very comprehensive training was performed in most cases with the iterative training process stopping only when a CER of 1\% or below was reached.
\end{itemize}

\subsubsection{Overall Time Expenditure and OCR Accuracy}

In this first experiment we evaluate the two main criteria for a workflow with considerable human interaction: the time that had to be invested both for obtaining a sufficient result regarding segmentation and OCR as well as the achieved OCR accuracy.
These criteria are heavily influenced by several factors which have to be taken into consideration.
First, the experience of the user:
An experienced user can be expected to be more efficient, both during the segmentation and the GT production phase.
In our experiments we differentiated between two groups of users:
On the one hand, there were several first time users with no experience with OCR4all and no or next to no experience with other OCR related tools and processes.
On the other hand, we had users with a solid general OCR background and an extensive history of using OCR4all, LAREX, and various transcription tools.
In the following we assign the labels 1 and 2 to the users of the respective groups, with a higher number indicating a more experienced user.
To distinguish the individual users from each group we assign additional labels (A, B, ...).
With the exception of one experienced user (digital humanities) all users are classical humanities scholars.
Before starting to work on their respective books on their own, all participants were introduced to the tool by one of the experienced users.

Second, challenges due to the book:
Different books can vary considerably, mainly regarding the number of pages, the complexity of the layout but also the print or scan quality and overall state of preservation.
Since the books utilized during our experiments did not show large discrepancies concerning the latter criteria we just provide the number of pages and distinct semantic layout classes.
Table \ref{tab:precise} sums up the results.

\paragraph{Results}


\begin{table}[!htb]
\centering
\caption{
Results of the precise segmentation and trained OCR of early printed books.
The \textit{Book}s are grouped by the experience of the processing \textit{User}s, first the unexperienced (\textit{1}), then the experienced ones (\textit{2}).
Regarding the \textit{Segmentation} we provide the number of pages (\textit{\#P}) and semantic region types (\textit{\#R}) that had to be distinguished as well as the time required for the entire book (\textit{$t_\text{Seg.}$}) and per page (\textit{$t_\text{Seg.}/P$}) on average.
For the OCR we indicate the maximum number of GT lines \textit{\#L} which was used to train the final OCR model along with the achieved CER.
Furthermore, the time required to correct all lines required to train the final model and one line on average is shown (\textit{$t_\text{Corr.}$}).
Finally, some \textit{Key Figures} are derived to ensure comparability of the works.
The overall manual time expenditure is calculated for the entire book (\textit{$t_\text{All}$}) by adding up the overall time required for segmentation and OCR and for an average single page (\textit{$t_\text{All} / P$}) by dividing by the number of pages.
For both user groups averaged values of all books (\textit{Mean}) and the corresponding standard deviation (\textit{StdDev.}) are provided if sensible.
}
\label{tab:precise}
\begin{tabular}{lc|rcrc|rcrr|rc}
\toprule
\textbf{Book} & \textbf{User} & \multicolumn{4}{c|}{\textbf{Segmentation}} & \multicolumn{4}{c|}{\textbf{OCR}} & \multicolumn{2}{c}{\textbf{Key Figures}}\\
Short & Exp. & \#P   &\#R& $t_\text{Seg.}$& $t_\text{Seg./P}$ & \#L   & CER   & $t_\text{Corr.}$ & $t_\text{Corr.}/L$ & $t_\text{All}$ & ${t_\text{All}/P}$ \\
& & & & [min] & [min] & & [\%] & [min] & [s] & [min] & [min] \\
\midrule
C1532a & 1A &  55   & 7 &  90 & 1.6 &   829 & 0.47  & 280 & 20 & 370 & 6.7  \\
C1532b & 1A & 130   & 7 & 110 & 0.8 &   611 & 0.73  & 146 & 14 & 256 & 2.0  \\
C1533  & 1A &  57   & 5 &  82 & 1.4 &   806 & 0.20  & 129 & 10 & 211 & 3.7  \\
C1535  & 1A &  96   & 7 & 104 & 1.1 &   723 & 0.39  & 176 & 15 & 280 & 2.9  \\
C1552  & 1A & 180   & 6 & 110 & 0.6 &   384 & 0.20  &  44 &  7 & 154 & 0.9  \\
C1554  & 1B &  81   & 6 &  66 & 0.8 &   487 & 0.36  &  76 &  9 & 142 & 1.8  \\
C1557  & 1B & 168   & 5 & 194 & 1.2 & 1,342 & 0.34  & 187 &  8 & 381 & 2.3  \\
C1558  & 1A &  94   & 8 & 139 & 1.5 &   751 & 0.25  & 183 & 15 & 322 & 3.4  \\
C1561  & 1B & 344   & 5 & 275 & 0.8 &   395 & 0.40  &  48 &  7 & 323 & 0.9  \\
C1563  & 1C & 158   & 5 & 140 & 0.9 & 1,175 & 0.60  &  95 &  5 & 235 & 1.5  \\
C1566  & 1D & 471   & 7 & 370 & 0.8 &   596 & 0.61  &  48 &  5 & 418 & 0.9  \\
C1568  & 1B & 342   & 5 & 223 & 0.7 &   406 & 0.24  &  36 &  5 & 259 & 0.8  \\
N1494  & 1E & 156   & 7 & 210 & 1.3 & 2,302 & 0.69  & 315 &  8 & 525 & 3.4  \\
N1494  & 1F & 157   & 7 & 360 & 2.3 &   969 & 0.82  &  97 &  6 & 457 & 2.9  \\
N1549  & 1G & 328   & 7 & 210 & 0.6 & 2,824 & 0.45  & 155 &  3 & 365 & 1.1  \\
P1474  & 1H & 198   & 4 &  29 & 0.1 &   700 & 0.90  & 230 & 20 & 259 & 1.3  \\
P1509  & 1I & 218   & 5 & 390 & 1.8 & 1,501 & 0.42  & 310 & 12 & 700 & 3.2 \\

\midrule

\textbf{Mean} & & \textbf{190} & \textbf{6.1} & & \textbf{1.1} & \textbf{988} & \textbf{0.47} & & \textbf{10} & & \textbf{2.3}  \\
\textbf{StdDev.} & & & & & \textbf{0.5} & & \textbf{0.22} & & \textbf{5.2} & & \textbf{1.5}  \\

\midrule

C1541  & 2B & 439   & 8 & 345 & 0.8 &   847 & 0.92  &  82 &  6 & 447 & 1.0   \\
C1566  & 2A & 240   & 7 &  80 & 0.3 &   599 & 0.57  &  45 &  4 & 118 & 0.5   \\
C1583  & 2A & 606   & 7 & 200 & 0.3 & 1,647 & 1.00  & 123 &  5 & 323 & 0.5   \\
C1594  & 2A & 420   & 8 & 200 & 0.5 &   352 & 0.50  &  26 &  4 & 226 & 0.5   \\
C1598  & 2B & 344   & 8 & 245 & 0.7 &   256 & 0.45  &  28 &  7 & 273 & 0.8   \\
N1498  & 2A & 161   & 6 & 130 & 0.8 &   622 & 0.30  &  22 &  2 & 152 & 0.9   \\
N1499  & 2A & 166   & 7 & 105 & 0.6 &   632 & 0.12  & 110 & 10 & 215 & 1.3   \\
N1506  & 2A & 215   & 8 & 180 & 0.8 & 3,161 & 0.20  &   - &  - &   - &   -   \\
P1484  & 2B & 372   & 3 &  65 & 0.2 &   226 & 0.34  &  22 &  6 &  80 & 0.2   \\

\midrule

\textbf{Mean} & & \textbf{329} & \textbf{6.9} & & \textbf{0.6} & \textbf{927} & \textbf{0.49} & & \textbf{5.5} & & \textbf{0.7}  \\
\textbf{StdDev.} & & & & & \textbf{0.2} & & \textbf{0.30} & & \textbf{2.4} & & \textbf{0.4}  \\

\bottomrule
\end{tabular}
\end{table}

For the less experienced users an average segmentation expense of slightly more than one minute per page was recorded with considerable variations among different users.
Fortunately, more experienced users can speed up the process considerably, resulting in about 36 seconds per page on average again with considerable variations depending on the layout complexity of the book.
Regarding the time expenditure required for correcting OCR results for GT production the vast majority of users invest less than ten seconds per line on average.
Both user groups achieved almost identical CERs (0.47\% and 0.49\%) by utilizing a very similar amount of GT (988 and 927 lines).
These results enable us to compare the time expenditure of the users on a more general level by taking the achieved OCR quality out of the equation.
Calculating the time required to process a book, both segmenting it and creating enough GT to obtain an average CER of below 0.5\% resulted in just 0.7 minutes per page for the experienced users.
Compared to the 2.3 minutes achieved by the unexperienced users this represents a speedup of more than factor 3.

\paragraph{Interpretation}

The times accounted for segmentation clearly show that performing a precise and fine-grained semantic segmentation of early printed books, even when using a comfortable and versatile tool like LAREX, can still amount to several hours of work for a single book.
On top comes the time to generate ground truth (either from scratch or by correcting an OCR result) and to train an OCR model.
Our experiments show that the total processing time from beginning (image processing, page segmentation) to end (OCR text with less than 0.5\% on average) is less than a day for books containing a few hundred pages. 

While this is still a far cry from an expectation of ``press a button, wait a few seconds, receive the results'', a meaningful comparison would be to look at the current practice of manual transcription as a baseline.
This is usually done either by a person sitting in front of two displays, one of which shows the page image of a book and the other a word processor.
The scholar alternately looks at the image, tries to decipher its text and enters it into the word processor on the other display.
The process is cumbersome, error prone (hence the practice of double keyboarding with two teams entering the same texts which will later be compared to spot transcription errors), and very time consuming.
The transcription of a whole book of several hundred pages can easily consume a few weeks. 
We did not thoroughly evaluate the manual transcription from scratch but to get a rough impression the users 2A and 2B transcribed a small number of pages from their respective books (C1541, P1484, and 1499).
Extrapolating the effort for the entire book led to an overall time expenditure of 44 hours for C1541, 47 hours for P1484, and 130 hours for N1499.
Our method therefore reduces the working time from a few weeks to a day, plus the additional effort to weed out the remaining OCR errors.

Next, the results indicate a high fluctuation of efficiency even within the two user groups, especially among the unexperienced users.
Out of the eight books which took longer than one minute per page, four were processed by the same user (1A).
The results of N1494 are especially eye-catching since the segmentation took the second user (1F) over 75\% longer than the first one (1E) despite both of them working on almost identical material.

Extremely complex layouts like the ones of N1498, N1499, and especially N1506 can be very challenging and not trivial to process even for very experienced users.
Having said that, in our experience these three examples are about as complex as it gets for early printed books, especially combined with our very strict and detailed segmentation guidelines.
Almost on the other end of the spectrum are books like P1484 where most pages are almost trivial to segment and therefore only require a minimal amount of time (around 10 seconds when processed by an experienced user).

Regarding the OCR correction it is noteworthy that four out of six books which required more than ten seconds per line were processed by a single user, the same that also achieved most of the slow segmentation results (1A).
Since there are no obvious reasons for this effect regarding the material we assume that some users simply require more time during the correction process maybe because they are too frightened to miss something.
This is also reflected by the general correction strategy of the two groups.
While the experienced users tend to simply scan the results by hopping between the line image and the OCR result on a word to word basis, the unexperienced users often first read the entire text in the line image and the OCR result separately, before performing a third check where smaller junks of the line are compared.
It is worth mentioning that cross checks of the produced GT showed no noteworthy effects regarding the quality of the transcriptions among different users.



Not only because of the fact that it was the most experienced user (2A) who achieved the worst OCR results of all books, we have no reason to believe that the user has a noteworthy influence on the OCR accuracy.
Most importantly, the reachable CER depends on the book and the contained typography as well as the amount of GT used for training.
The obtained results underline this assumption almost perfectly.


While the discussed key figures are very helpful to obtain an overall impression of the amount of manual effort required to process early printed books with OCR4all, further experiments are required to get a deeper understanding of the effects of the iterative training approach and the influence of segmentation guidelines.

\subsubsection{Evaluating the Iterative Training Approach}

\begin{figure}[!htb]
    \centering
    \includegraphics[width=0.8\linewidth]{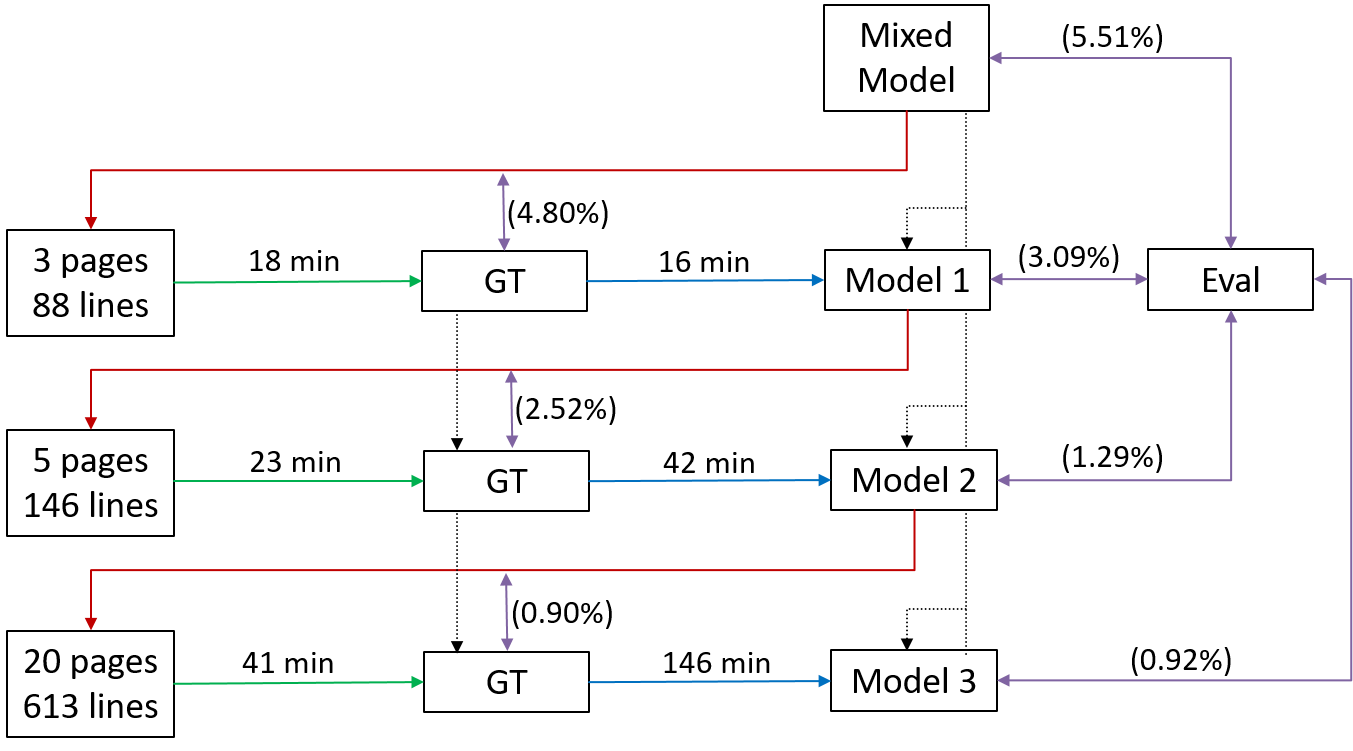}
    \caption{
    Schematic representation of the iterative training approach and its evaluation.
    As an example we used book C1541 processed by an experienced user.
    For comparison we refer to the first line of Table \ref{tab:eval_it-train}.
    To begin with, the user selects a few pages (here 3 pages comprising 88 lines) and applies a suitable \textit{Mixed Model} to it.
    After investing 18 minutes to correct the results a first evaluation shows that the \textit{Mixed Model} achieved a CER of 4.80\% on the first batch of lines.
    Next, the produced GT can be used to train a first book-specific model (\textit{Model 1}) which required 16 minutes, using the initial \textit{Mixed Model} as a starting point.
    \textit{Model 1} is then applied to the next batch (5 pages / 146 lines).
    After correcting the erroneous results (23 minutes, 2.52\% CER) a second book-specific model is trained (\textit{Model 2}, 42 minutes) using all available GT (8 pages / 234 lines) and again building from the initial \textit{Mixed Model}.
    This process is repeated until a satisfactory CER is reached or the entire book is transcribed.
    For evaluation purposes a separate \textit{Eval} dataset can be utilized which was not part of any training set.
    By applying the \textit{Mixed Model} and the models produced during each iteration to this dataset and evaluating the results we can compare the models objectively.
    }
    \label{fig:eval_schema}
\end{figure}

The manual correction effort not only scales with the number of lines that have to be corrected but also with their recognition quality.
To be able to thoroughly evaluate the effects and benefits of the iterative training many different values and results were recorded.
Since their evaluation and interpretation is a quite complex task we first introduce them and describe them in detail in Figure \ref{fig:eval_schema} before we list the results of selected works in Table \ref{tab:eval_it-train}.

\paragraph{Results}

\begin{table}[!htb]
\centering
\caption{
Evaluation of the iterative training approach.
For each \textit{Book} processed by a \textit{User} we provide all values and results necessary to reconstruct and evaluate the progress of the GT production and training.
In the \textit{New Data} column the number of the newly added pages (\textit{\#P}) and the corresponding number of lines (\textit{\#L}) is listed, as well as the time required to produce the transcription.
Furthermore, the CER is given which is calculated from the OCR result achieved by the model from the previous iteration and the newly created GT.
For comparison, the CER achieved on the separate and constant evaluation set (\textit{Eval}) is recorded.
\textit{All Data} shows the number of available GT pages and lines at this point which then serve as training data for the new model which is used for the next iteration.
In the \textit{Correction} columns we compare the actual required correction time when applying the iterative training approach (\textit{ITA}) with the projected time when only using the output of the mixed model (\textit{MM}) to get to the point where the final (and in case of \textit{MM} the first) model is trained. The speed up factor (\textit{SU}) is calculated for each book for the two user groups separately and for both groups combined.
}
\label{tab:eval_it-train}
\begin{tabular}{cc|c|ccccc|c|cc|ccc}
\toprule
\textbf{Book} & \textbf{User} & \textbf{It.} & \multicolumn{5}{c|}{\textbf{New Data}} & \textbf{Eval} & \multicolumn{2}{c|}{\textbf{All Data}} & \multicolumn{3}{c}{\textbf{Correction}} \\
Short & Exp. & & \#P & \#L & $t_\text{Corr.}$ & $t_\text{Corr.}$ & CER & CER & \#P & \#L & ITA & MM & SU  \\
& & & & & [min] & [s/L] & [\%] & [\%] & & & [min] & [min] & \\
\midrule
C1541 & 2B & 1 &  3 &    88 &  18 &   12 &  4.80 & 5.51 &  3 &   88 \\
      &    & 2 &  5 &   146 &  23 &  9.5 &  2.52 & 3.09 &  8 &  234 \\
      &    & 3 & 20 &   613 &  41 &  4.0 &  0.90 & 1.29 & 28 &  847 \\
      &    & 4 &  - &     - &   - &    - &     - & 0.92 &  - &    - & 82 & 169 & 2.1 \\
      
\midrule

P1484 & 2B & 1 &  5 &   110 &  14 & 7.6 &  3.53 &  3.95 &  5 &   110  \\
      &    & 2 &  6 &   116 &   8 & 4.1 &  0.89 &  1.48 & 11 &   226  \\
      &    & 3 &  - &     - &   - &   - &     - &  0.34 &  - &    - & 22 & 29 & 1.3 \\

\midrule

N1499 & 2A & 1 &  2 &   105 &  65 &  37 & 25.22 & 23.59 &  2 & 105  \\
      &    & 2 &  3 &   138 &  20 & 8.7 &  0.54 &  2.23 &  5 & 243  \\
      &    & 3 &  5 &   389 &  25 & 3.9 &  1.24 &  1.63 & 10 & 632  \\
      &    & 4 &  - &     - &   - &   - &     - &  0.20 &  - &   - & 110 & 474 & 3.6  \\
      
\midrule
 & & & & & & & & & & & \multicolumn{2}{r}{\textbf{Mean(2):}} & \textbf{2.3} \\
\midrule

C1557 & 2B & 1 &  4 &   104 &  16 & 9.2 &  2.00 & 10.00 &  4 &   104 \\
      &    & 2 & 11 &   307 &  70 &  14 &  6.06 &  8.64 & 15 &   411 \\
      &    & 3 & 15 &   407 &  56 & 8.3 &  1.60 &  1.17 & 30 &   818 \\
      &    & 4 & 20 &   524 &  45 & 5.2 &  0.26 &  0.65 & 50 & 1,342 \\
      &    & 5 &  - &     - &   - &   - &     - &  0.34 &  - &     - & 187 & 206 & 1.1 \\

\midrule

C1558 & 1A & 1 &  4 &   125 &  38 &  18 & 15.31 & 16.86 &  4 &   125 \\
      &    & 2 &  8 &   251 &  60 &  14 &  1.28 &  0.65 & 12 &   376 \\
      &    & 3 & 12 &   375 &  85 &  14 &  0.58 &  0.34 & 24 &   751 \\
      &    & 4 &  - &     - &   - &   - &     - &  0.25 &  - &     - & 183 & 225 & 1.2 \\

\midrule

C1566 & 1D & 1 &  5 &   122 &  15 & 7.4 &  3.85 &  4.27 &  5 &   122 \\
      &    & 2 &  6 &   126 &  18 & 8.6 &  3.15 &  1.45 & 11 &   248 \\
      &    & 3 & 12 &   348 &  15 & 2.6 &  0.22 &  0.99 & 23 &   596 \\
      &    & 4 &  - &     - &   - &   - &     - &  0.61 &  - &     - & 48 & 74 & 1.5 \\

\midrule
 & & & & & & & & & & & \multicolumn{2}{r}{\textbf{Mean(1):}} & \textbf{1.3} \\
\midrule
 & & & & & & & & & & & \multicolumn{2}{r}{\textbf{Mean:}} & \textbf{1.9} \\
\bottomrule
\end{tabular}
\end{table}

There are several interesting things to be taken away from the results summarized in Table \ref{tab:eval_it-train}.
First of all, it is shown that the iterative training approach yields a significant speedup regarding the correction time.
On average the manual effort is almost cut in half (average speedup factor 1.9, last column) with the experienced users benefitting considerably more compared to the unexperienced ones (factor 2.3 and 1.3).

Another eye-catching abnormality are the discrepancies between the performances of the same models on the new and the eval data.
While some deviations had to be expected and can be considered negligible others seem to be too substantial to be disregarded as variance.
For example, when processing C1557 achieves a good CER of 2\% on the new data but at the same time struggles severely with the eval data (10\% CER). An explanation is given in the next section.

\paragraph{Interpretations}

\begin{figure}[!htb]
    \centering
    \includegraphics[width=\linewidth]{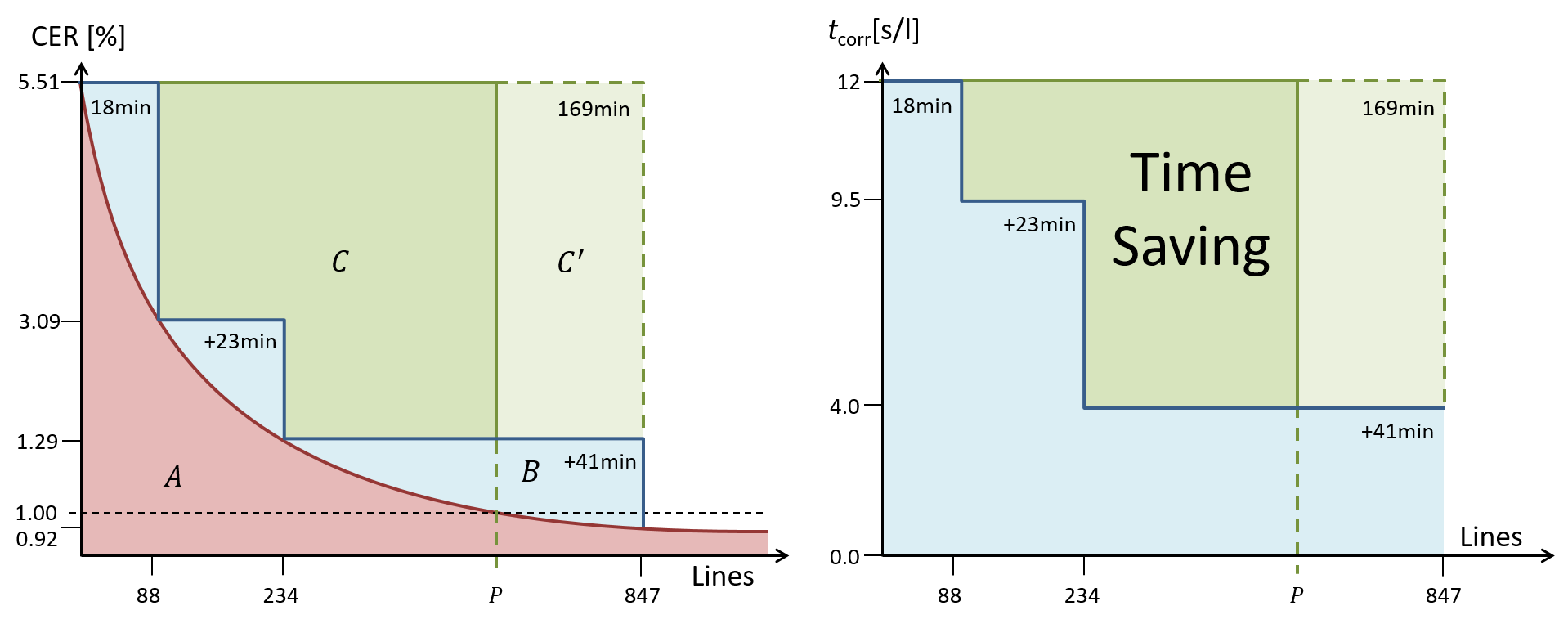}
    \caption{Left: Analysis of the iterative training approach using C1541.
    The goal is to reach point \emph{P} which represents the (unkown) number of lines necessary to reach a CER of below 1\%.
    In theory, the red curve describes the unknown relation between GT lines used for training and the achieved CER and therefore represents the theoretical ideal iterative training approach that produces and applies a new OCR model after the transcription of a single new line of GT.
    The real route chosen by user 2B is shown by the blue stair case function.
    Green depicts a single training approach using the same final number of line of GT as user 2B ($C+C'$) and the perfect but unknown number of lines ($C$).\\
    Right: This time the $t_\text{corr}$-coordinate shows the manual effort necessary to correct a single (which highly depends on the CER).
    The time saved by the iterative training approach (in case of user 2B and theoretically ideal) is represented by the green area minus the blue area.
    In this case this represents a reduction of the manual effort by 51\% which equates to a speedup of 2.1.
    }
    \label{fig:ita}
\end{figure}

Admittedly, the projection of the speedup achieved by the iterative training approach is quite rough since the factor depends a lot on the pages the mixed model was applied to, which is also shown by the high fluctuations among the speedup factors.
Moreover, in a real-world application scenario there has to be some kind of training and testing during the correction phase in order to know when to stop as the results from Table \ref{tab:precise} have shown that the number of lines needed to reach a certain CER varies considerably.
Figure \ref{fig:ita} depicts this problem and graphically explains the gain obtained by the iterative training approach.

Determining the ideal training route is no trivial task and depends on several factors.
To begin with, the user has to estimate how many lines are necessary to reach the desired CER.
Due to the variety of the material this is very challenging, even for experienced users, resulting in over and under estimations of the required amount.
The smaller the chosen steps are the more accurate the convergence to the optimal value $P$ (Figure \ref{fig:ita}) becomes.
One (theoretical) approach is training a model each time a new line of GT is added (red curve, left), however this is not sensible.
The other end of the spectrum is represented by correcting the output of the mixed model until the presumably required number of GT lines is reached (green), which discards the gain of correcting lines with an improving CER (area ratio on the right).
Consequently, the optimal or rather a sufficient real world solution has to lie somewhere in between these two extremes.
The available hardware plays an important role as it directly influences the training duration.
For example, most training processes can be completed within a couple of minutes when using several GPUs, allowing the user to continue the transcription almost instantly.
When no GPU support is available a training can take several hours, requiring the user to perform different tasks.

Despite the complexity of optimizing the iterative training approach, its general benefits are clear and the results confirm the expectations.
This speedup is due to the fact that the average correction time per line clearly correlates with the quality (CER) of the underlying OCR result.
The only exception can be seen in the iterations 1 and 2 of book C1566 where it took the user even a little longer to correct a line, despite starting from a somewhat better recognition result.
Naturally, comparisons like this are only viable for a single user and within the same book.
Since a visual inspection of the concerned pages did not lead to new insights it stands to reason that human factors like tiredness, form on the day, etc. play a non-negligible role.
Furthermore, it is worth mentioning that the correction time will not decrease linearly since the user will always require a certain amount of time for grasping the line image and reading the OCR text even with a theoretical CER of 0.0\%.

In case of C1557 the reason for the striking difference between the CER on the new and the eval data is that most of the Camerarius books incorporate a frequent change of two completely different Antiqua fonts: upright and italics.
Of course, other books make use of different fonts as well but mostly in less frequent layout parts like headings while in case of Camerarius they are used for the main text and therefore often fill entire pages.
So when the ratio of pages/lines printed in upright and in italics varies considerably between the new and the eval data, this of course also effects the obtainable CER.
In the case of C1557 the new data in the first iteration did not contain any italics lines so the mixed model which was mainly trained on Antiqua upright performed quite well.
On the contrary, the eval data had a significant portion of italics lines the default model could not handle.
After creating some GT of italics lines during the first iteration, which is indicated by the relatively high CER (6\%), the newly trained model can deal with both fonts, resulting in a considerably better CER on the eval set.
To counter this problem we trained a new mixed model, after several Camerarius books had been processed, which was not only to deal considerably better with the upright/italics problem but also with the Greek characters.

N1499 is another interesting case.
First, despite printed in a rather normal looking Bastarda font and being among the best books in the corpus in terms of image quality, the CER obtained from applying a mixed model is by far the worst that occurred during our experiments, yielding a CER of around 25\%.
Naturally, this also explains the unusually high time expenditure necessary for correction we mentioned during the discussion of Table \ref{tab:precise}, since the correction of an OCR result flawed to that degree is a very cumbersome task and barely faster than transcribing the line from scratch, if at all.
Second, while the recognition quality on the new data in the second iteration is great (0.54\%) the resulting model performs significantly worse on the new data of the next iteration (1.24\%).
While the processing user (2A, the most experienced user participating in our experiments) could neither explain this by the scan quality or the use of different fonts, a look at the confusion tables quickly identified the problem:
In iteration 2, the error distribution looked like the ones produced by a well converged model, mainly consisting of the misrecognition of rare capital letters or typical OCR error like the confusion of \emph{c} and \emph{e}.
However, in iteration 3 the most frequent errors were dominated by previously negligible errors which all related to the characters \emph{x}, \emph{v}, and \emph{j}.
It turned out that the printer decided to use printing types with considerably different looking glyphs for different kinds of marginalia.

While some marginalia serve as reminders for the reader or simply repeat some keywords from the main text, others are referencing other books using Roman numerals which are numbered and therefore rely on the affected characters a lot.
The effect was further fortified as during the third iteration the pages added by the user contained plenty of marginalia containing references.
This phenomenon did not only affect the recognition quality but also the correction time.
Despite the considerably higher CER in iteration 3 the user handled these lines more than twice as fast on average as the lines from iteration 2.
While the effect can partially be explained by the accumulation of very short lines (mainly marginalia), another reason, according to the processing user, clearly was the shift in the error distribution.
As explained before, the main cause of error are the highly flawed reference numberings while the remainder of the text was recognized with very high accuracy.
Understandably, this effects the amount of effort required for correction heavily since the errors are often clumped and can often be corrected very efficiently, for example by deleting an entire number and simply typing in \emph{xxviij} without even having to use the virtual keyboard once.

Concerning the training duration (machine time without human intervention) we do not want to go into detail in this paper as the required times considerably depend on many factors including the available hardware, the amount of available GT, many training parameters, especially the use of data augmentation, and the activation of early stopping.
In our experience a modern PC or laptop is enough to quickly perform standard training runs within one to two hours while even extensive book-specific training processes can be completed over night.
During the course of our experiments we set up an instance of OCR4all on a server where the data could also be accessed by a highly performant GPU cluster allowing to complete most of the training processes in a couple of minutes.

\subsubsection{Segmentation Without Semantic Classification}

\begin{figure}[!htb]
    \centering
    \includegraphics[width=\linewidth]{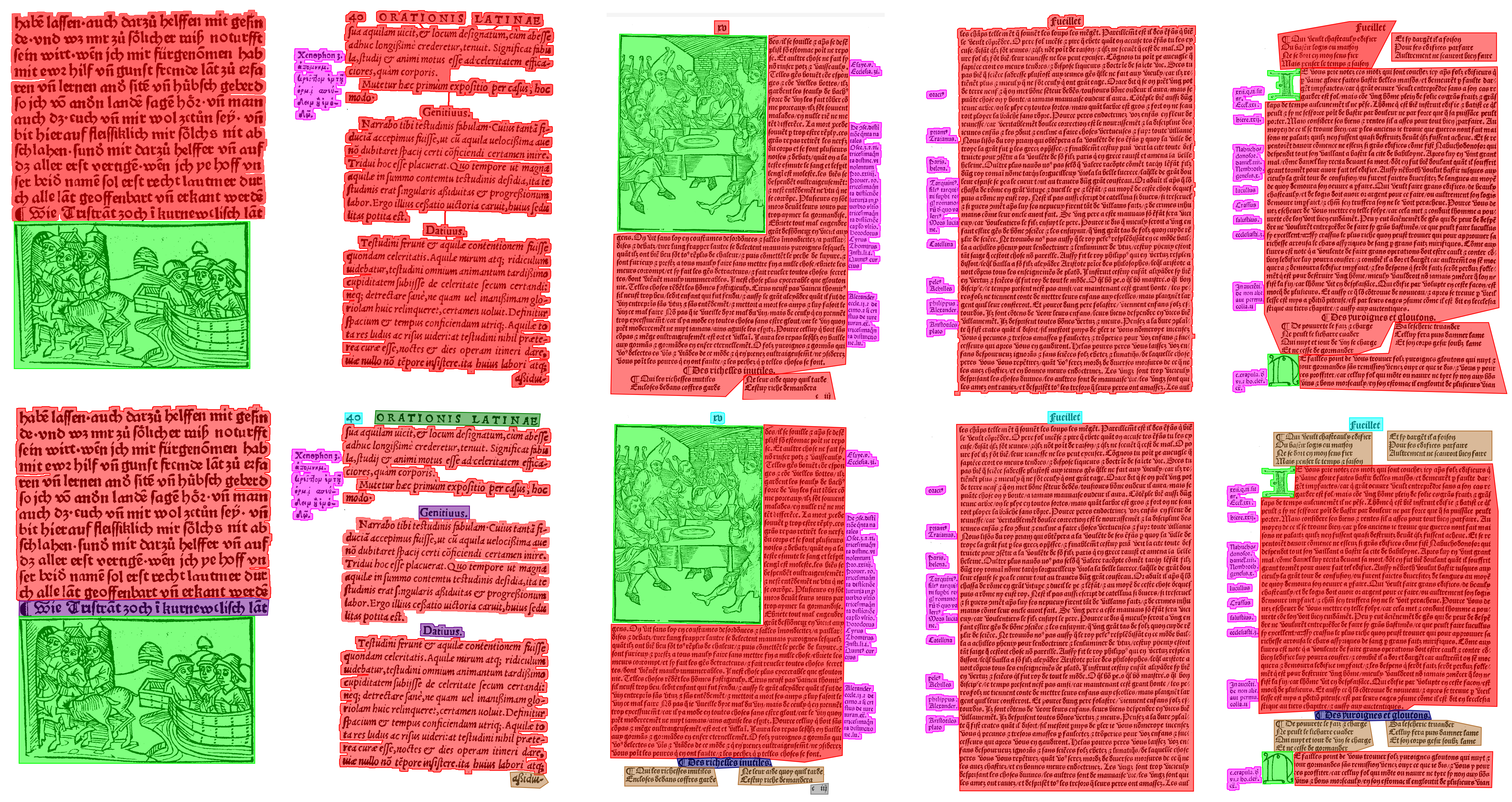}
    \caption{
Representative example image of three books showing the difference between the basic (top row) and exact (bottom row) segmentation approach.
From left to right: P1484, C1541, N1506 (three images).
    }
    \label{fig:seg_basic-ext}
\end{figure}

As our first experiment has shown, the segmentation step can be considerably more time consuming than the OCR, even when aiming for very low CERs.
Of course this is especially true for voluminous books since the effort necessary to obtain a certain recognition accuracy does not scale with the size of the book, but the segmentation effort does.
However, the required manual work to segment a book can be severely cut down when the aspirations regarding semantic classification are less strict.
Therefore, we conducted another experiment where the single goal of the segmentation is to provide the subsequent OCR with the means sufficient to produce the required output.
Apart from a clean text/non-text separation this also includes ensuring the correct reading order.
Figure \ref{fig:seg_basic-ext} shows the desired results for example pages of the three books we used for this experiment: P1484, C1541, and N1506.

We selected these books due to their widely differing layout properties:
While P1484's very simple layout apart from images and running text only uses a single additional semantic element (chapter headings), C1541 and N1506 both make use of a wide variety.
Still, also these two books differ considerably since C1541's layout elements for the most part are simply arranged from top to bottom in a one column layout, with the comparatively rarely used marginalia being the only exception.
N1506 however not only has plenty of marginalia but also incorporates a complex two column sub layout on every other page.

For our experiment the users 1C and 2B both segmented 20 representative pages of each the books twice, first using the basic approach (text/non-text and reading order only) and then the complex approach from the first experiments including a fine-grained semantic classification.

\paragraph{Results}

For all sub task we recorded the required times which are presented in Table \ref{tab:seg_basic-exact}.

\begin{table}[!htb]
\centering
\caption{
Comparison of two users (1C and 2B) and two different segmentation approaches regarding the segmentation with LAREX.
Apart from the processed \textit{Book} and the experience of the \textit{User} we differentiate between the two segmentation \textit{Mode}s (see Figure \ref{fig:seg_basic-ext}) basic (\textit{B}) and exact (\textit{E}).
We give the \textit{Time}s required for the segmentation of 20 representative pages and for a single page on average.
Finally we calculate the ratio of the basic approach to the exact one (\textit{B / E}) and between the less experienced user and the more experienced one (\textit{1C / 2B}).
}

\label{tab:seg_basic-exact}
\begin{tabular}{cc|c|cc|cc}
\toprule
\textbf{Book} & \textbf{User} & \textbf{Mode} & \multicolumn{2}{c|}{\textbf{Time}} & \textbf{B / E} & \textbf{1C / 2B}\\
& & & [s] & [s / P] & & \\
\midrule
P1484       & 2B & B &   120 &  6 & 0.75  & 3.04 \\
            & 2B & E &   160 &  8 &       & 3.84 \\
            & 1C & B &   365 & 18 & 0.59  &  \\
            & 1C & E &   615 & 31 &       &  \\
\midrule
C1541       & 2B & B &   255 & 13 & 0.46  & 3.43 \\
            & 2B & E &   560 & 28 &       & 3.41 \\
            & 1C & B &   875 & 44 & 0.46  &  \\
            & 1C & E & 1,920 & 96 &       &  \\
\midrule
N1506       & 2B & B &   805 & 40 & 0.66  & 1.49 \\
            & 2B & E & 1,220 & 61 &       & 1.61 \\
            & 1C & B & 1,200 & 60 & 0.61  &  \\
            & 1C & E & 1,965 & 98 &       &  \\
\bottomrule
\end{tabular}
\end{table}

As expected the basic segmentation approach requires considerably less time than the exact one, leading to an average savings of 38\% for the experienced and 45\% for the unexperienced user.
Regarding the comparison of two users with a different degree of experience the results in general show the expected tendency, namely a much faster processing by the more experienced user.
On average it takes the novice user 2.65 times longer to perform the basic segmentation on the three books.
This factor even rises slightly to 2.95 when an exact segmentation is required.

\paragraph{Interpretations}

When taking a closer look at the individual books it is eye-catching that the most time can be saved using the basic segmentation approach with C1541.
This makes sense since dropping the requirement of semantic classification reduces the complexity of C1541 and thereby the typical Camerarius books significantly, especially because the reading order in the single column layout is always automatically correct, as soon as the marginalia, if present, have been cut off.
In comparison P1484 does not allow to save as much time since the required manual effort remains the same for many pages as they simply contain running text only or running text with images which both are either segmented correctly automatically or have to be manually corrected, regardless of the segmentation mode.
Consequently, there is a difference in the results from the (semi-)manual segmentation of the headings.
As for N1506, bigger savings were not reached because the complex layout requires most of the manual interventions in order to ensure the reading order and not because of a correct semantic segmentation.
For example, when the marginalia are correctly separated from the main text LAREX automatically assigns the correct type.
In the two column sections the user can save some time by not separating the headings but the main matter of expense, that is the column separation in itself, remains.

It is worth mentioning the discrepancy between the two users, while being very similar for P1484 and C1541, is comparatively low for N1506, which at first seems odd, since N1506 is the book with the most challenging layout not only in this experiment but in the entire corpus.
This effect can be explained when looking at the advantages an experienced has over an unexperienced one.
The three main aspects are:

\begin{itemize}
    \item Mental understanding of the layout.
    \item Optimized application of advanced segmentation techniques.
    \item Raw technical ability.
\end{itemize}

We think that experienced users can understand a layout considerably faster and then act accordingly right away.
Furthermore, they can use LAREX much more efficiently by putting its automatic features to work.
However, when dealing with extremely complex layouts that require lots of manual interventions, the overall degree of advantage is dominated by the effect of raw technical ability, which we consider to be significant but not as high as the one of the other advantages.

\subsection{Fully Automated Processing}

Next, we reduce the manual effort to a minimum by choosing a fully automated approach.
Since the Dummy Segmentation of OCR4all relies on the layout analysis functionality comprised in the OCRopus 1 line segmentation step, more complex layouts cannot expected to be solved without human intervention.
Consequently, we first focus our in-detail evaluation on the 19\textsuperscript{th} century Fraktur models whose layout is usually rather trivial, before we turn to more challenging material.

\subsubsection{19th Century Novels}

This experiment was performed on ten German Fraktur novels from the corpus described above in \ref{tab:fraktur-novels} using the Calamari Fraktur 19\textsuperscript{th} century ensemble which was trained on a wide variety of data also derived from 19\textsuperscript{th} century Fraktur novels (see \cite{reul2019state} for details).

We randomly selected ten pages from each novel and processed them fully automatically with OCR4all as well as with ABBYY Finereader Engine CLI for Linux\footnote{\url{https://www.ocr4linux.com/en}} version 11 together with ABBYY's historical Fraktur (Gothic) module and Old German language settings.
The results were compared by calculating the CER on a page to page basis.
To ensure a fair comparison several regularizations, for example the normalization of the long and short version of the \emph{s}, were performed beforehand.

\paragraph{Results}

Table \ref{tab:abbyy-ocr4all} sums up the results.
The values show that OCR4all considerably outperforms ABBYY Finereader on every single book resulting in an average improvement of over 84\% and a relative improvement of almost 8 with respect to the error rate.
On eight of the ten books CERs of below 1\% are achieved while six books even yielded error rates below 0.5\%.
Wild fluctuations in CER can be observed for ABBYY Finereader (best: 0.48\%, worst: 27\%) but also for OCR4all (best: 0.06\%, worst: 4.89\%) caused by the highly variant quality of the scans as shown in Figure \ref{fig:fraktur_imgs}.

\begin{table}[!htb]
\centering
\caption{CERs achieved by \textit{ABBYY} Finereader and \textit{OCR4all} when being applied fully automatically to different \textit{Book}s.
The final two columns indicate the percent error reduction \textit{ErrRed.} and the improvement factor \textit{Impr.} improvements yielded by OCR4all over ABBYY.
Furthermore, we provide the average (\textit{Avg.}) over all books for each value.
}

\label{tab:abbyy-ocr4all}
\begin{tabular}{ccccc}
\toprule
\textbf{Book} & \textbf{ABBYY} & \textbf{OCR4all} & \textbf{ErrRed.} & \textbf{Impr.}\\
\midrule
F1781 & 2.9  & 0.60 & 79.3 &  4.8 \\
F1803 &  27  & 4.89 & 81.9 &  5.5\\
F1810 & 3.8  & 0.61 & 84.0 &  6.2\\
F1818 &  10  & 1.35 & 86.6 &  7.5\\
F1826 & 1.1  & 0.06 & 94.4 &  18\\
F1848 & 0.93 & 0.20 & 78.5 &  4.7\\
F1851 & 1.0  & 0.16 & 84.0 &  6.3\\
F1855 & 4.0  & 0.33 & 91.8 &  12\\
F1865 & 1.6  & 0.18 & 88.8 &  8.9\\
F1870 & 0.48 & 0.13 & 72.9 &  3.7\\
\midrule
\textbf{Avg.} & \textbf{5.3} & \textbf{0.85} & \textbf{84.2} & \textbf{7.8} \\

\bottomrule
\end{tabular}
\end{table}

\paragraph{Interpretations}

ABBYY struggles noticeably with substantially soiled pages, recognizing lines in regions showing dirt or bleed through on a regular basis, resulting in gibberish OCR output.
OCR4all shows only few segmentation errors, with the main problem being left out page numbers, which happens due to a heuristic in the OCRopus 1 line segmentation script that ignores lines that contain less than three CCs.

\begin{table}[!htb]
\centering
\caption{
The ten most common confusions over all ten books for \textit{ABBYY} Finereader and \textit{OCR4all}, consisting of the \textit{GT}, the prediction (\textit{OCR}), the counted number of occurrences (\textit{CNT}) and the percent contribution (\textit{\%}) of a given confusion to the overall number of errors.
Whitespaces are shown as \textit{␣} and empty cells denote no prediction.}
\label{tab:error_analysis}
\begin{tabular}{ccrr|ccrr}
\toprule
\multicolumn{4}{c|}{\textbf{ABBYY}} & \multicolumn{4}{c}{\textbf{OCR4all}}\\
\midrule
\textbf{GT} & \textbf{OCR}  & \textbf{CNT} & \textbf{\%} & \textbf{GT} & \textbf{OCR} & \textbf{CNT}  & \textbf{\%}\\
\midrule
␣ &   & 64 & 2.6 &       ␣ &   & 63 & 11.9\\
  & ␣ & 57 & 2.3 &       n & u & 14 & 2.7\\
s & S & 57 & 2.3 &      f & s &  12 & 2.3\\
  & , & 50 & 2.0 &      i & l &  12 & 2.3\\
e & c & 40 & 1.6 &      r & t &  12 & 2.3\\
e & " & 40 & 1.6 &      " & , &   9 & 1.7\\
s & r & 40 & 1.6 &      , &   &   9 & 1.7\\
- & " & 39 & 1.6 &      i &   &   8 & 1.5\\
  & . & 36 & 1.5 &      c & e &   8 & 1.5\\
x & " & 35 & 1.4 &        & ,&    6 & 1.1\\

\midrule
\multicolumn{3}{r}{Remaining} &81.6 & \multicolumn{3}{r}{Remaining} & 71.1\\
\bottomrule
\end{tabular}
\end{table}

First of all it is apparent that the error distribution of the results produced by OCR4all is more top heavy, with the top 10 making up for almost 30\% of the total errors, compared to the one of ABBYY (less than 20\%).
However, a closer look shows that the distributions are actually quite similar to each other, apart from the top error of OCR4all, namely the deletion of whitespaces, which is responsible for almost 12\% of the errors alone.
Interestingly, while insertions and deletions of whitespaces represent the top two errors for ABBYY and OCR4all also fails to predict them on a regular basis, insertions of whitespaces do not occur in the top ten of OCR4all at all.
The remainder of the most frequent OCR4all errors looks as expected, containing well-known errors like the confusion of similar looking (at least in 19\textsuperscript{th} Fraktur script) characters like \emph{n} and \emph{u}, \emph{f} and \emph{s} (originally predicted as the long s and then regularized), or \emph{c} and \emph{e} as well as the insertions and deletions of tiny elements like commata, sometimes also as part of quotation marks.
ABBYY seems to struggle with quotations marks as well but mostly by confusing seemingly random glyphs like \emph{e}, \emph{-}, and \emph{x} with them.
Even when considering that the original recognition mostly showed french quotation marks (\emph{»«}) which might explain some of the confusions the notable accumulation of these errors remains inexplicable.
Furthermore, some of the aforementioned typical OCR errors like the confusions of \emph{e} and \emph{c} and \emph{s} and \emph{S} still are surprising since one would expect the powerful dictionary and language modelling capabilities of ABBYY to deal with these errors quite comfortable.
A possible explanation is that these postcorrection operations do not change characters that have been recognized with a certain degree of confidence to prevent the introduction of errors when ``improving'' out of dictionary words like unusual proper names.
It is noticeable that 27 of the 57 \emph{s/S} errors appear in a single book (F1851) but closer inspection did not lead to any new insights as the used glyphs differ considerably, are often recognized correctly, and no pattern regarding the misrecognitions could be observed.

Again, it has to be emphasized that these very low CERs can only be achieved when a highly performant mixed model is available.
In this case we were able to rely on a strong voting ensemble perfectly fitting the evaluation material.
Unfortunately, comparable ensembles are not available for other scripts and languages, yet.

\subsubsection{Early Printed Books}
As expected, the fully automatic processing of early printed books is a tricky task and its applicability highly depends on the layout and typography of the book at hand.
The results presented above as well as some additional experiments led to the following mostly qualitative observations:

\begin{itemize}
    \item The current setup can deal with relatively simple layouts consisting of a single or several well separated columns quite reliably.
    When several columns have to be identified the user needs to specify the maximum number of columns occurring on a page.
    \item Despite the lack of an explicit text/non-text segmentation, the combination of OCRopus 1 line segmentation and Calamari's recognition module is surprisingly robust against non-text elements like noise, artistic border elements, images, and swash capitals.
    Even if parts of these elements make it into a text line they often do not deteriorate the text recognition result since Calamari will ignore them due to the lack of a confident recognition of available characters.
    \item Marginalia which are located very close to the main text often cannot get separated correctly, leading to significant errors in the reading order.
    \item Treating a page that comprises highly varying font sizes, for example a very prominent heading line and many running text lines whose characters are not even half as high, as a single text segment can lead to wrongly segmented lines.
    This happens because the line segmentation estimates the most likely height of a line on page level and then tries to find fitting lines.
    A preceding region segmentation prevents this problem from occurring.
    \item The available mixed models work reasonably well on the majority of books achieving an average CER of 7.7\% on the corpus we used for our evaluations (see Table \ref{tab:early_printed_books}).
    However, since this is probably not good enough for most use cases, book-specific training is necessary.
    Additionally, the CERs vary considerably, ranging from below 2\% to over 25\% on the new GT of the first iteration of each book, underlining the problematic of mixed models we discussed before.
\end{itemize}

%% file: chapters/05_discussion.tex
\section{Discussion}
\label{sec:discussion}

Before discussing the results and their meaning in detail we sum up the main findings of our experiments:

\begin{itemize}
    \item When dealing with challenging early printed books unexperienced users had to invest 2.3 minutes per page on average to perform a precise segmentation and to reach a CER of below 0.5\% which highlights the effectiveness of the proposed approach.
    Experienced users can perform much more efficiently reaching a speedup factor of more than 3 (0.7 minutes per page on average).
    \item The iterative training approach yielded significant speedups (factor 1.9) compared to the naive correction of the output of the mixed model.
    \item A basic segmentation approach which only ensures a sufficient text/non-text separation and a correct reading order reduces the manual effort required for segmentation considerably by a factor of more than 2.5.
    \item The experiments with 19\textsuperscript{th} century Fraktur novels showed that a fully automated application of OCR4all is not only possible but can be highly precise on material with moderate complex layouts and if a suitable OCR model is available (average CER of 0.85\% compared to ABBYY's 5.3\%).
\end{itemize}

The obtained results show that OCR4all fulfills its purpose by enabling non-technical users to capture even the earliest printed books completely by their own and with great quality despite the challenges provided by complex layout and irregular typography.
Due to our strict demands regarding the semantic classification of layout elements and our goal of high OCR quality, a considerable amount of manual work was required and accepted.
While the experiments showed that even non-technical users without any background or previous experience in OCR were comfortably able to successfully work with OCR4all, the results also showed that there is a learning curve and experience is key.
This holds true for both the segmentation as well as the OCR, with experienced users being almost twice as fast when it comes to segmenting a page or transcribing a text line compared to unexperienced users on average.
However, the quality of the result was not influenced by the experience of the user, with both groups achieving an excellent average CER of slightly below 0.5\%.

Regarding the two main steps which require manual intervention, segmentation and OCR, the first one seems to show more room for improvement since the OCR of historical printings made great progress over the last few year which could also be observed during our experiments.
Calamari's training and recognition capabilities combined with the easy to use iterative training approach provided by OCR4all allow the users to utilize state-of-the-art deep learning software and accuracy improving techniques like pretraining, voting, and data augmentation without ever being forced to acquire a deeper understanding of the technical concepts behind them.
As shown by the evaluations, CERs below 1\% or even 0.5\% should almost be considered the norm after a thorough book-specific training was performed.
The segmentation using LAREX proved to be intuitive and highly accurate.

While the usage of OCR4all reduces the manual effort necessary to transcribe an early printed books tremendously, especially compared to the fully manual approach which often required several weeks of full-time transcribing to process a single book, we still think that there is room for improvement.

While a fully automated approach where it is not necessary to look at every single page seems to be currently out of reach, not only due to the complexity of the layouts but also due to the very high demands of the users regarding the quality of the segmentation and degree of semantic distinction, we think that the efficiency of this part of the workflow can be further increased.
Especially the segmentation of the Camerarius books appears to be a rather frustrating task since the one-column layout in itself is rather trivial but the repeated manual classification of semantic sub regions can be exhaustive.
Pushing the labeling part to a later stage in the workflow, and for example perform it after the line segmentation or even after the OCR, seems to be a viable solution but the realization is not trivial.
While it is possible to first capture the entire text and then add semantic labels later, for example by encoding them using TEI, the positional information with regards to the scan would get lost, considerably limiting the possibilities regarding the presentation of the result.
Having said that, the region coordinates do not necessarily have to be recorded during the region segmentation step.
An alternative could be to first perform a less time consuming text/image separation like the basic segmentation approach we evaluated earlier.
Next, the user could perform the line segmentation and apply the iterative training approach.
Finally, the semantic classification of layout elements would take place at the very end of the workflow by making use of the line coordinates.
For example, when dealing with a sub heading with adjacent running text above and below, the user could simply select the line and apply the new type, resulting in three new regions.
Especially when processing works with similar layout properties like the Camerarius books from our experiments this approach would significantly speed up the segmentation and labeling process without noteworthy affecting the rest of the workflow.

When discussing the means necessary to increase the degree of automation the need for a book-specific training also plays an important role.
While our experiments have confirmed the effectiveness and efficiency of the iterative training approach it still represents a time-consuming task.
As thoroughly argued during the first two sections of this work the need for book-specific training highly depends on the material at hand but most importantly on the intended use of the results.
In this paper we mainly focused on books which are captured within projects that intend to produce quotable texts as their final result and therefore need to be corrected to their full extent.
Under these circumstances it has to be considered negligent not to perform a book-specific training, since the ongoing transcription/correction of the book produces comprehensive amounts of GT anyway.
Nevertheless, we are aware of the fact that there are other projects and use cases that, despite dealing with early printed books, aim for a more quantitative approach and therefore are willing to make sacrifices regarding text quality and semantic labeling.
Our evaluations showed that a basic segmentation approach that simply ensures a proper text/non-text separation and a correct reading order can save around 40\% of the time required for segmentation.
While this represents a substantial speedup the required manual effort is probably still too high for some areas of application.
Yet, we have seen that the segmentation approaches currently available in OCR4all cannot deal with more complex layouts in a (close to) fully automatic manner.

Even after obtaining a sufficient segmentation result with minimal or no human effort, the OCR represents the next big challenge.
When aiming for the greatest extent of automation book-specific training does not seem like a viable solution.
Yet, as we have seen above, despite our repertoire mixed models trained on a wide variety of fonts geared towards the recognition of different font classes, a great or at least sufficient OCR quality cannot be guaranteed at all when working with early printed books.
Of course, a wider variety of more specialized mixed models can help to improve results.

Suitable grouping criteria besides the general font type and the age of the books could be the region or even the printing shop a book was printed in.
However, producing GT to train these models is a very cumbersome and time consuming, especially when starting from scratch (or from very general mixed models), like we did with Camerarius.
A possible solution would be to try to make use of already available data as much as possible, to identify appropriate font clusters, to train the corresponding mixed models and to apply them to the parts of a text which have been identified to contain the respective font.
First steps in this direction have been made by Weichselbaumer et al. \cite{weichsel2019font} who make use of the \textit{Gesamtkatalog der Wiegendrucke}\footnote{\url{https://www.gesamtkatalogderwiegendrucke.de/}} (GW) and its side project the \textit{Typenrepertorium der Wiegendrucke}\footnote{\url{https://tw.staatsbibliothek-berlin.de/}} (TW) that store extensive information concerning the typography used during the 15\textsuperscript{th} century and use a CNN that is presented 25 random crops from a page in order to classify the main font.
While the use of the glyphs indexed in GW and TW is definitely promising the font recognition on a page to page basis does not seem ideal as a more fine-grained classification on word or even sub-word level would allow a more precise application of suitable OCR models.
As we have seen during our Camerarius use case, the use of several fonts or even scripts on a single page or even within a line is not uncommon.
Therefore, we think that a line-based approach like the one we proposed for the automatic semantic tagging of a historical lexicon \cite{reul2019automatic} is better suited for the task since the lines have to be identified anyway before the subsequent OCR step can take place.
We already showed Calamari's ability to reliably and accurately distinguish between different fonts even when their typefaces were very similar to each other.
Kraken has also taken a few first steps in that direction by providing a generic model that can differentiate between Latin, Arabic, Greek, and Syriac script.
After detecting the script suitable OCR models can be applied to the (parts of) lines which match.

In general, the applicability of fully automated methods does not only depend on the intended usage of the results but also on the material at hand.
We have shown that OCR4all can achieve excellent results on 19\textsuperscript{th} century Fraktur models considerably outperforming the commercial state-of-the-art tool ABBYY Finereader.
However, these results could only be obtained because of the relatively straight forward layout and the availability of a suitable mixed OCR model, which are both aspects which cannot always be considered to be the case even for 19\textsuperscript{th} century material or even newer. If there is either a complex layout or no good recognition model available the result will be far from excellent for any OCR engine. If the layout is either simple or can be ignored because the segmentation is treated separately the OCR engine with the better recognition engine will win. While OCR4all delivered superior results compared to ABBYY on 19\textsuperscript{th} century novels, this was not a universal ranking: On issues of the German periodical ``Daheim'', a German journal published between 1864 and 1943, ABBYY recognized text segments with a CER of 0.07\% outperforming Calamari with its standard voting ensemble (0.09\%). Apparently, ABBYY's internal recognition model for Fraktur had been trained on a very similar font.

As an example regarding the availability of suitably mixed models for more modern printings our efforts to support the COST\footnote{\url{https://www.cost.eu/cost-actions/}}-funded project \textit{Distant Reading for European Literary History}\footnote{\url{https://www.distant-reading.net/}} come to mind where we had to deal with novels mostly printed around 1900 in Antiqua.
While the layout was trivial and the print quality and state of preservation was comparatively excellent our available models could not deal with the typography or rather the languages.
Since among others the novels were printed in Portuguese, Romanian, and Hungarian many glyphs and diacritica were unknown, resulting in CERs of around 10\% when applying the Antiqua modern standard model.
While a few pages of GT were sufficient to train highly performant (CER < 0.5\%) book-specific models this underlines the problematic nature of having suitably mixed OCR models available does not only relate to early printed books but also to comparatively modern material, if hitherto unknown glyphs are encountered. In this case we had to add example lines with the missing characters to our training corpus to train a new mixed model.

\begin{figure}[!htb]
    \centering
    \includegraphics[width=0.65\linewidth]{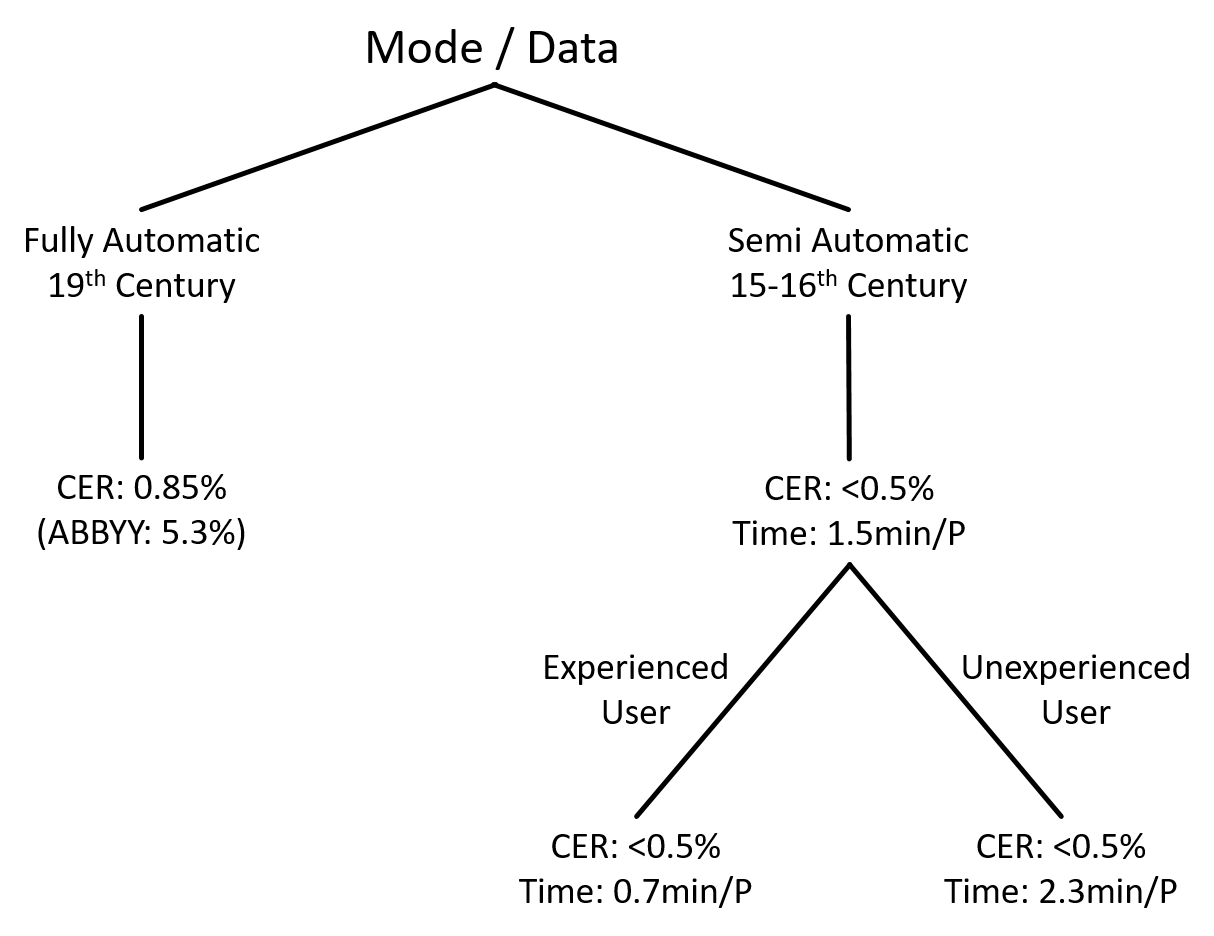}
    \caption{Final summary of the results dependent on the material and the processing user.}
    \label{fig:results-graph}
\end{figure}

To sum up, despite the open questions and challenges demonstrated above, OCR4all can become a cornerstone when it comes to the high quality capturing of historical printings (a final summary of the obtained results is shown in Figure \ref{fig:results-graph}).
By reducing the required technical know how to a minimum it is now possible for humanities scholars to take the acquisition of their much desired and needed textual research data into their own hands.

%% file: chapters/06_conclusion.tex
\section{Future Work}
\label{sec:conclusion}

In this section we first discuss features we would like to integrate into OCR4all or its submodules in the future before concluding the paper by giving an outlook on the general future of OCR4all.

First of all, one of the main goals to address is the obvious lack of a more potent segmentation method that can deal with (more) complex layouts in a highly automized way.
Therefore, we aim to include a trainable pixel classifier in order to either provide a valid starting point for other segmentation approaches by classifying pixels and consequently connected contours as text, image, and noise or even perform a fine-grained semantic markup \cite{wick2018fully}.
Of course, a more powerful segmentation approach must also comprise a more sophisticated method for the determination of the reading order which also has to be integrated into LAREX.
To generate the reading order, the idea is to allow the user to comfortably specify rules based on the detected region types as well as their absolute and relative position.

Regarding training and recognition we want to provide the user with the possibility to comfortably train several type-specific models for a single work.
This can be very helpful, when the book comprises a few, possibly highly differing, fonts or even scripts, e.g.\ Latin and Greek.
A book-specific trained script detection model in fact should be expected to perform considerably better, not to mention that it is not feasible to train generic mixed models to recognize book-specific fonts.
Consequently, an appropriate GT production and training functionality will have to be supported within OCR4all.

Furthermore, in order to train more robust models, a more flexible selection of lines for training, recognition, and correction is desirable as this allows to train models using GT that is widely spread over the course of the entire book.
This would help to further optimize the iterative training approach by integrating active learning, i.e.\ adding lines to the existing GT pool, which the current models had problems recognizing, instead of random ones.
The effectiveness of this approach has already been shown by \cite{reul2018improving} who were able to considerably improve the OCR results (average gain of 16\%, maximum gain of 32\%) by purposefully adding lines to the training set, that showed the largest disagreement between the separate outputs of the voters.

Apart from the voting, there are several other use cases in which we want to profit from the character confidences provided by Calamari.
First, the correction process can be supported by highlighting suspicious characters.
Second, by averaging the confidence values over several lines it is possible to identify segments or pages which contain a worse recognition result compared to the rest of the book.
This could help to identify text parts that suffer from an increased amount of degradation, contain segmentation errors, use a different type of font, etc.
Third, the average confidence calculated over a representative number of recognized lines can serve as a form of quality estimation.
We know that the confidence values correlate with the recognition rate and that the neural networks tend to overestimate their performance.
Therefore, we hope that it is possible to use a lot of existing measurements to derive a model which is able to estimate the true recognition accuracy based on the average confidence \cite{springmann2016automatic}.
In addition to the automatic selection of the best fitting model for given data, this would be particularly helpful when the goal of a OCR process is to reach a certain recognition quality (for example 2\% CER are sufficient for most NLP tasks) and it is unclear whether the output of a mixed model suffices or if book-specific training is required.

A most desirable issue that has to be addressed as soon as possible is the incorporation of a post processing step, for example using dictionaries or language modelling.
However, especially for (very) early printed books this is not a trivial task due to the lack of consistent spelling rules and the frequent use of abbreviations.

Moreover, the scan preparation, which as of now is either skipped or performed externally by using Scan Tailor, should be integrated directly into the web GUI.

In order to allow non-technical users to keep all semantic annotations without having to deal with the peculiarities of the PageXML format, one or several alternatives to the two existing output formats are imperative.
Since TEI is considered the go-to format for textual markup applications a comfortable and attractive solution would be to allow the user to simply specify at least some basic mappings between PageXML types and TEI tags and then let OCR4all export the result as a valid TEI file.

Another useful feature would be an integrated fuzzy search that allows the user to spot keywords or phrases even if they contain OCR errors.
The basis for this is already laid because of Calamari's extended recognition output which does not only provide the final textual OCR result but also additional information like the alternatives for each character and their respective confidence values.

A particularly interesting and challenging goal is to overcome the additional difficulties of handwritten recognition.
Despite the general workflow being very similar, there are several steps that would require adaptations, for example the line segmentation step since handwritten text often does not consist of single characters, which the current line segmentation approach heavily relies on, but features joined-up writing.
Yet, there are already a few pertinent open-source algorithms available, for example \cite{gruning2018two}, which can at least serve as a valid starting point.
Apart from the line segmentation the remaining steps work quite similarly to when dealing with printed texts.
Actually, OCR4all has already been successfully applied to a Greek manuscript (Aëtius Amidenus - Libri medicinales, 16\textsuperscript{th} century), achieving character recognition rates in the mid nineties when using only a few hundred lines of GT.

As mentioned above OCR4all's primary field of application was planned to be the local setup at a single users desktop PC or laptop.
However, with some manageable extensions regarding a project and user administration system as well as an interface to a resource scheduling manager, OCR4all can be deployed and run as a full-featured web service.
This would be especially helpful for institutions or working groups who want to share their resources among themselves in order to work collaboratively.
Even without further extensions a collaborative approach is already possible:
During our experiments we set up an instance for several users to cooperate in a somewhat coordinated way which proofed to be highly effective.

A key aspect remains the optimization of the teaching material associated with the tool.
In the future we want to build from the already existing written guides not only by adding screencasts or even tutorial videos but by setting up a knowledge base, most likely in the form of a Semantic MediaWiki, that not only contains the guides mentioned above in a more modular form but also extends them with and crosslinks them to the theoretical concepts behind each individual step of the OCR4all workflow. 
Combined with a public repository for GT and models, best practices as well as an assembly of frequently occurring difficulties and proven ways to successfully deal with them, this would provide the community with a place to share material, knowledge, problems, and solutions.

Despite these comprehensive plans for the future, we already reached our main goal of creating a tool which provides non-technical users with access to a powerful and easy to use OCR workflow.
This is not only shown by the evaluations but also by the successful application in numerous real-world projects where OCR4all leads to significant speedups of the OCR of our precious cultural heritage.